\documentclass[10pt,journal,compsoc]{IEEEtran}
\usepackage[pagebackref=false,breaklinks=true,letterpaper=true,allcolors=blue,colorlinks,bookmarks=false]{hyperref}

\ifCLASSOPTIONcompsoc
  \usepackage[nocompress]{cite}
\else
  \usepackage{cite}
\fi
\ifCLASSINFOpdf
  \usepackage[pdftex]{graphicx}
\else
  \usepackage[dvips]{graphicx}
\fi
\usepackage[cmex10]{amsmath}
\usepackage{algorithm}
\usepackage{algorithmic}
\usepackage{amssymb}
\usepackage{enumitem}
\usepackage{sidecap}
\usepackage{multirow}

\def\eg{e.g.\ }
\def\etc{etc.\ }
\def\cf{cf.\ }
\def\Eg{E.g.\ }
\def\etal{et al.\ }
\def\ie{i.e.\ }
\def\vs{vs.\ }
\def\wrt{w.r.t.\ }
\def\R{\mathbb{R}}
\def\N{\mathbb{N}}
\def\I{\mathcal{I}}

\def\M{\mathcal{M}}
\def\T{\mathcal{T}}
\def\w{\textbf{w}}
\def\x{\textbf{x}}
\def\z{\textbf{z}}
\def\l{\boldsymbol{\ell}}
\def\a{\boldsymbol{\alpha}}
\def\Order{\mathcal{O}}
\def\train{{train} }
\def\test{{test} }
\def\validation{{validation} }

\begin{document}
%
\title{Expanded Parts Model for Semantic Description of Humans 
in Still Images}
%
%
%
%

\author{Gaurav~Sharma,~\IEEEmembership{Member,~IEEE},
        Fr\'ed\'eric~Jurie,
        and~Cordelia~Schmid,~\IEEEmembership{Fellow,~IEEE}
\IEEEcompsocitemizethanks{\IEEEcompsocthanksitem Gaurav Sharma is with Max Planck Institute for 
Informatics, Germany. The majority of the work was done when he was with GREYC CNRS UMR 6072, Universit\'e
de Caen Basse-Normandie, and INRIA-LEAR, France. http://www.grvsharma.com
\IEEEcompsocthanksitem Fr\'ed\'eric Jurie is with GREYC CNRS UMR 6072,
Universit\'e of Caen Basse-Normandie, France. http://jurie.users.greyc.fr
\IEEEcompsocthanksitem Cordelia Schmid is with LEAR, INRIA Grenoble Rhone Alpes, France.
http://lear.inrialpes.fr
}
\thanks{}}

%
%

\markboth{}%
{Sharma \MakeLowercase{\textit{et al.}}: Expanded Parts Model}
%



\IEEEtitleabstractindextext{%
\begin{abstract}
We introduce an Expanded Parts Model (EPM) for recognizing human attributes (\eg young, short hair,
wearing suits) and actions (\eg running, jumping) in still images. An EPM is a collection of part
templates which are learnt discriminatively to explain specific scale-space regions in the images
(in human centric coordinates). This is in contrast to current models which consist of a relatively
few (\ie a mixture of) `average' templates. EPM uses only a subset of the parts to score an image
and scores the image sparsely in space, \ie it ignores redundant and random background in an image.
To learn our model, we propose an algorithm which automatically mines parts and learns corresponding
discriminative templates together with their respective locations from a large number of candidate
parts. We validate our method on three recent challenging datasets of human attributes and actions.
We obtain convincing qualitative and state-of-the-art quantitative results on the three datasets.
\end{abstract}

}

\maketitle

\IEEEdisplaynontitleabstractindextext

%
\IEEEpeerreviewmaketitle

\IEEEraisesectionheading{\section{Introduction}\label{sec:introduction}}

\IEEEPARstart{T}{he} focus of this paper is on semantically describing humans in still images using
attributes and actions. It is natural to describe a person with attributes, \eg age, gender,
clothes, as well as with the action the person is performing, \eg standing, running, playing a
sport. We are thus interested in predicting such attributes and actions for human centric still
images. While actions are usually dynamic, many of them are recognizable from a single static image,
mostly due to the presence of (i) typical poses, like in the case of running and jumping, or (ii) a
combination of pose, clothes and objects, like in the case of playing tennis or swimming. 

With the incredibly fast growth of human centric data, \eg on photo sharing and social networking
websites or from surveillance cameras, analysis of humans in images is more important than ever. The
capability to recognize human attributes and actions in still images could be used for numerous
related applications, \eg indexing and retrieving humans \wrt queries based on higher level semantic
descriptions. 

Human attributes and action recognition have been addressed mainly by (i) estimation of human pose
\cite{YangCVPR2010, YaoCVPR2010a} or (ii) with general non-human-specific image classification
methods \cite{DelaitreBMVC2010, SharmaBMVC2011,SharmaCVPR2012, YaoCVPR2011}. State-of-the-art action
recognition performance has been achieved without solving the problem of pose estimation
\cite{DelaitreBMVC2010, Everingham2011,SharmaCVPR2012, YangCVPR2010}, which is a challenging problem
in itself. Concurrently, methods have been proposed to model interactions between humans and the
object(s) associated with the actions \cite{DelaitreNIPS2011, DesaiCVPRW2010, GuptaPAMI2009,
PrestPAMI2011, YaoCVPR2010, YaoCVPR2010a}. In relevant cases, modelling interactions between humans
and contextual objects is an interesting problem, but here we explore the broader and complementary
approach of modeling appearance of humans and their immediate context  for attribute and action
recognition. When compared to methods exploiting human pose and human-object interactions, modelling
appearance remains useful and complementary, while it becomes indispensable in the numerous other
cases where there are no associated objects (\eg actions like running, walking) and/or the pose is
not immediately relevant (\eg attributes like long hair, wearing a tee-shirt).

\begin{figure}[t] 
\centering 
\includegraphics[width=\columnwidth, trim=5 420 455 0, clip]{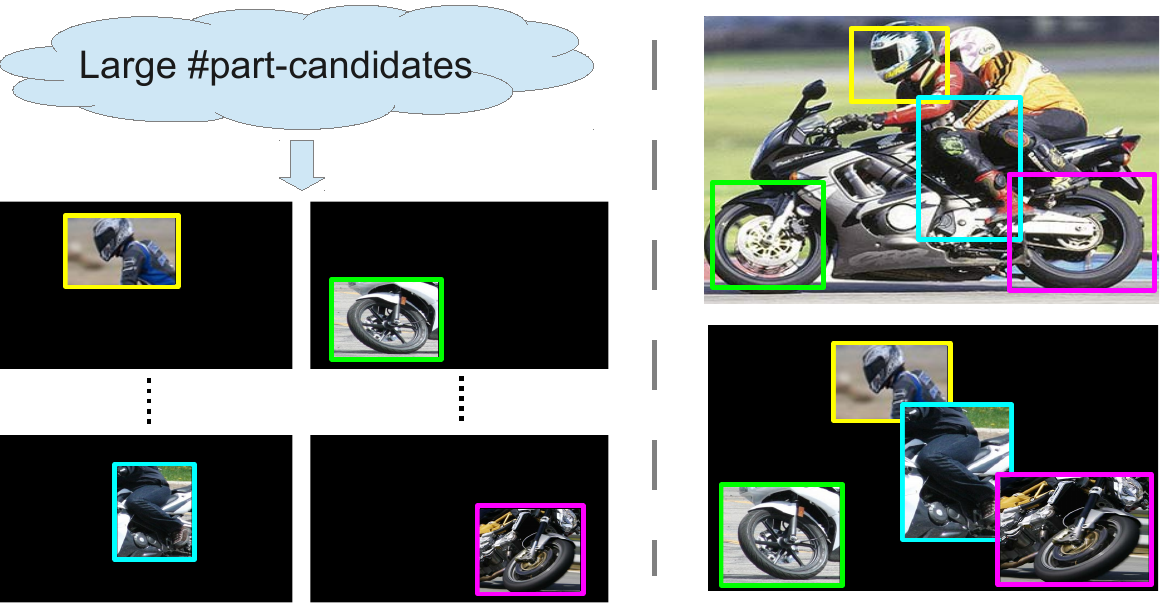} 
\vspace{-2em}
\caption{Illustration of the proposed method. During training (left) discriminative templates are
learnt from a large pool of randomly sampled part candidates. During testing (right), the most
relevant parts are used to score the test image.} \label{fig:epm_illus} 
\end{figure}

 \begin{figure*} 
\centering 
\includegraphics[width=0.85\textwidth, trim=2 420 234 0, clip]{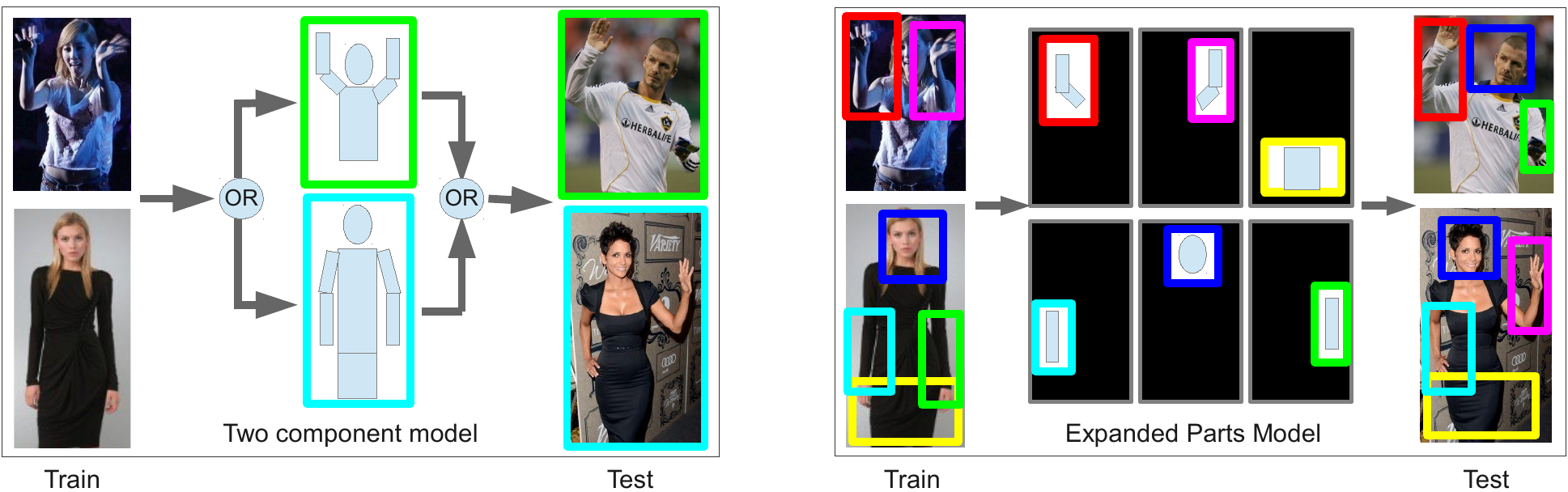} 
\vspace{-1em}
\caption{Illustration of a two-component model vs.\ the proposed Expanded Parts Model.  In a
component-based model (left) each training image contributes to the training of a single model and,
thus, its parts only score similar images. In contrast, the proposed EPM automatically mines
discriminative parts from all images and uses all parts during testing.  Also, while for
component-based models, only images with typical training variations can be scored reliably, in the
proposed EPM sub-articulations can be combined and score untypical variations not seen during
training.}
\label{fig:illus_dpm} 
\end{figure*}

In this paper, we introduce a novel model for the task of semantic description of humans, the
Expanded Parts Model (EPM). The input to an EPM is a human-centered image, \ie it is assumed that
the human positions in form of bounding boxes are available (\eg from a human detection algorithm).
An EPM is a collection of part templates, each of which can explain specific scale-space regions of
an image. Fig.~\ref{fig:epm_illus} illustrates learning and testing with EPM. In part based models
the choice of parts is critical; it is not immediately obvious what the parts might be and, in
particular, should they be the same as, or inspired by, the biologic/anatomic parts. Thus, the
proposed method does not make any assumptions on what the parts might be, but instead mines the
parts most relevant to the task, and jointly learns their discriminative templates, from among a
large set of randomly sampled (in scale and space) candidate parts. Given a test image, EPM
recognizes a certain action or attribute by scoring it with the corresponding learnt part templates.
As human attributes and actions are often localized in space, \eg shoulder regions for `wearing a
tank top', our model explains the images only partially with the most discriminative regions, as
illustrated in Fig.~\ref{fig:epm_illus} (right). During training we select sufficiently
discriminative spatial evidence and do not include regions with low discriminative value or regions
containing non-discriminative background. The parts in an EPM compete to explain an image, and
different parts might be used for different images. This is in contrast with traditional part based
discriminative models where all parts are used for every image.  

EPM is inspired by models exploiting sparsity. In their seminal paper, Olshausen and Field
\cite{OlshausenVR1997} argued for a sparse coding with an over-complete basis set, as a possible
computation model in the human visual system. Since then sparse coding has been applied to many
computer vision tasks, \eg image encoding for classification \cite{YangCVPR2009, YangECCV2010},
image denoising \cite{MairalJMLR2010}, image super-resolution \cite{YangTIP2010}, face recognition
\cite{WrightPAMI2009} and optical flow \cite{JiaICCV2011}. EPM employs sparsity in two related ways;
first the image scoring uses only a small subset of the model parts and second scoring happens with
only partially explaining the images spatially. The former model-sparsity is inspired by the coding
of information sparsely with an over-complete model, similar to Olshausen and Field's idea
\cite{OlshausenVR1997}. Owing to such sparsity, while the individual model part interactions are
linear, the overall model becomes non-linear \cite{OlshausenVR1997}.  The second spatial sparsity is
a result of the simple observation that many of the attributes and actions are spatially localized,
\eg for predicting if a person is wearing a tank top, only the region around the neck and shoulders
needs to be inspected, hence the model shouldn't waste capacity for explaining anything else (in the
image space).

\begin{figure*}
\centering 
\includegraphics[width=\textwidth, trim=0 338 60 0, clip]{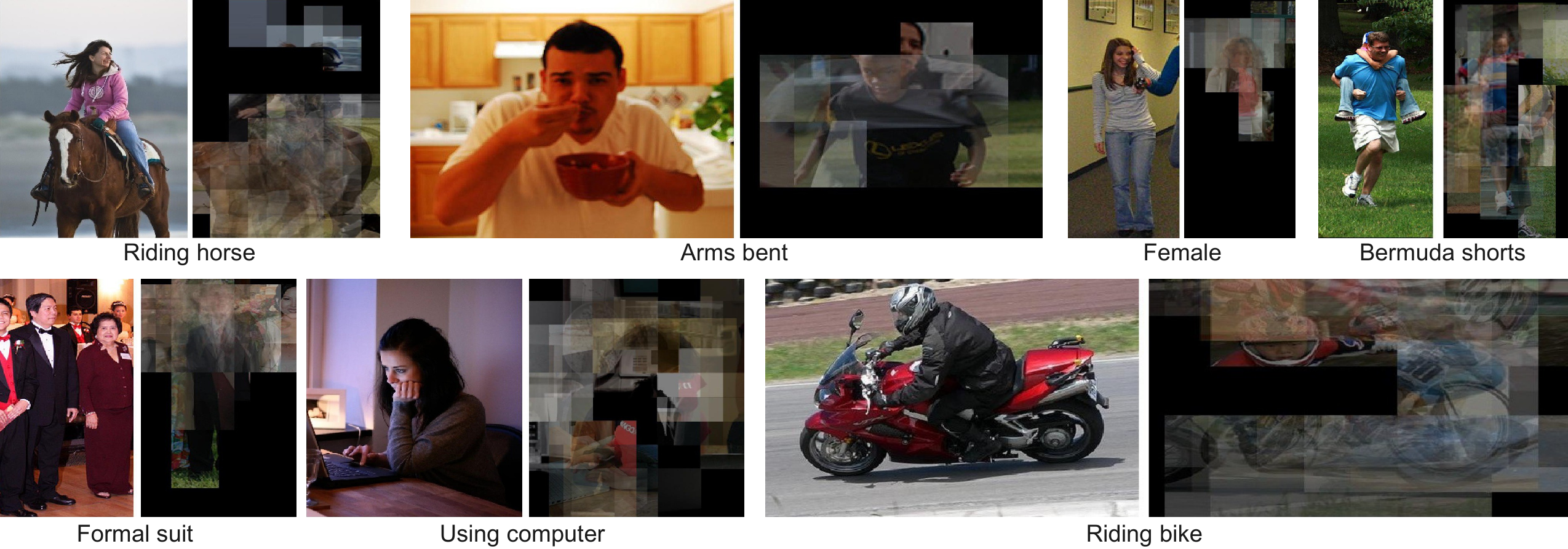} 
\vspace{-1.5em}
\caption{Illustrations of scoring for different images, for different attributes and actions. Note
how the model s  cores only the discriminative regions in the image while ignoring the
non-discriminative or background regions (in black). Such spatial sparsity is particularly
interesting when the discriminative information is expected to be localized in space like in the
case of many human attributes and actions.} 
\label{fig:illus_recons} 
\end{figure*}

To learn an EPM, we propose to use a learning algorithm based on regularized loss minimization and
margin maximization (Sec.~\ref{sec:approach}). The learning algorithm mines important parts for the
task, and learns their discriminative templates from a large pool of candidate parts. 

Specifically, EPM candidate parts are initialized with $\Order(10^5)$ randomly sampled regions from
training images. The learning then proceeds in a stochastic gradient descent framework
(Sec.~\ref{sec:solve}); randomly sampled training image is scored using up to $k$ model parts, and
the model is updated accordingly (Sec.~\ref{sec:scoring_func}). After some passes over the data, the
model is pruned by removing the parts which were never used to score any training image sampled so
far. The process is repeated for a fixed number of iterations to obtain the final trained EPM. The
proposed method is validated on three publicly available datasets of human attributes and actions,
obtaining interesting qualitative (Sec.~\ref{sec:exp_qual}) and greater than or comparable to
state-of-the-art quantitative results (Sec.~\ref{sec:exp_quant}). A preliminary version of this work
was reported in Sharma \etal \cite{SharmaCVPR2013}.

\section{Related work}

We now discuss the related work on modeling, in particular models without parts, part-based
structured models and part-based loosely structured models. 

\subsection{Models without parts} 
Image classification algorithms have been shown to be successful for the task of human action
recognition, see Everingham \etal \cite{Everingham2011} for an overview of many such methods. Such
methods generally learn a discriminative model for each class. For example, in the Spatial Pyramid
method (SPM), Lazebnik \etal \cite{LazebnikCVPR2006} represent images as a concatenation of
bag-of-features (BoF) histograms \cite{CsurkaSLCV2004, SivicICCV2003}, with pooling at multiple
spatial scales over a learnt codebook of local features, like the Scale Invariant Feature Transform
(SIFT) of Lowe \cite{LoweIJCV2004}. Lazebnik \etal \cite{LazebnikCVPR2006} then learn a
discriminative class model $\w$ using a margin maximizing classifier, and score an image as
$\w^\top\x$, with $\x$ being the image vector. The use of histograms destroys `template' like
properties due to the loss of spatial information. Although SPM has never been viewed as a template
learning method, methods using gradients based features \cite{DalalCVPR2005, BenensonCVPR2012,
DollarPAMI2014, FelzenszwalbPAMI2010} have been presented as such, \eg the recent literature is full
of visualizations of templates (class models) learnt with HOG-like \cite{DalalCVPR2005} features,
\eg \cite{FelzenszwalbPAMI2010, PandeyICCV2011}. Both, SPM and HOG based, methods have been applied
to the task of  human analysis \cite{KhanIJCV2013,DelaitreBMVC2010}, where they were found to be
successful. We also formulate our model in a discriminative template learning framework. However, we
differ in that we learn a collection of templates instead of a single template.  

In the recently proposed Exemplar SVM (ESVM) work, Malisiewicz \etal \cite{MalisiewiczICCV2011}
propose to learn discriminative templates for each object instance of the training set independently
and then combine their calibrated outputs on test images as a post-processing step. In contrast, we
work at a part level and use all templates together during both training and testing. More recently,
Yan \etal \cite{YanECCV2012} proposed a 2-level approach for image representation. Similar to our
approach it involves sampling image regions, but while they vector quantize the region descriptors,
we propose a mechanism to select discriminative regions and build discriminative part based models
from them.  

Works have also been reported using features which exploit motion for recognizing and localizing
human actions in videos \cite{JainCVPR2013, JainCVPR2014, OneataCVPR2014,
WangICCV2013,LaptevCVPR2008, SimonyanNIPS2014}. Wang and Schmid \cite{WangICCV2013} use
trajectories, Jain \etal use tubelets \cite{JainCVPR2014} while Simonyan \etal
\cite{SimonyanNIPS2014} propose a two-stream convolutional network. Here, we are interested in human
action and attribute recognition, but only from still images and hence do not have motion
information.

\subsection{Part-based structured models}
Generative or discriminative part-based models (\eg the Constellation model by Fergus \etal
\cite{FergusIJCV2007} and the Discriminative Part-based Model (DPM) by Felzenszwalb \etal
\cite{FelzenszwalbPAMI2010}), have led to state-of-the-art results for objects that are rigid or, at
least, have a simple and stable structure. In contrast humans involved in actions can have huge
appearance variations due to appearance changes (\eg clothes, hair style, accessories) as well as
articulations and poses. Furthermore, their interaction with the context can be very complex.
Probably because of the high complexity of tasks involving humans, DPM does not perform better than
SPM for human action recognition as was shown by Delaitre \etal \cite{DelaitreBMVC2010}. Increasing
the model complexity, \eg by using a mixture of components \cite{FelzenszwalbPAMI2010}, has been
shown to be beneficial for object detection\footnote{See the results of different versions of the
DPM software http://people.cs.uchicago.edu/$\sim$rgb/latent/ which, along with other improvements,
steadily increase the number of components and parts.}.  Such increase in model complexity is even
more apparent in similar models for finer human analysis, \eg pose estimation \cite{DesaiECCV2012,
YangCVPR2011, ZhuCVPR2012}, where a relatively large number of components and parts are used. Note
that components account for coarse global changes in aspect/viewpoint, \eg full body frontal image,
full-body profile image, upper body frontal image and so on, whereas parts account for the local
variations of the articulations, \eg hands up or down.  Supported by a systematic empirical study,
Zhu \etal \cite{ZhuBMVC2012} recently recommended the design of carefully regularized richer (with a
larger number of parts and components) models.  Here, we propose a richer and higher capacity model,
but less structured, the Expanded Parts Model.

In mixture of components models, the training images are usually assigned to a single component (see
Fig.\ \ref{fig:illus_dpm} for an illustration) and thus contribute to training one of the templates
only. Such clustering like property limits their capability to generate novel articulations, as
sub-articulation in different components cannot be combined. Such clustering and averaging are a
form of regularization and involve manually setting the number of parts and components. In
comparison, the proposed EPM does not enforce similar averaging, nor does it forbid it by
definition. It can have a large number of parts (up to the order of the number of training images)
if found necessary despite sufficient regularization. Part-based deformable models initialize the
parts either with heuristics (\eg regions with high average energy \cite{FelzenszwalbPAMI2010}) or
use annotations \cite{DesaiECCV2012}, while EPM systematically  explores parts at a large number of
locations, scales and atomicities and selects the ones best suited for the task.

\subsection{Part-based loosely structured models} 
EPM bears some similarity with Poselets by Bourdev \etal 
\cite{BourdevICCV2011, BourdevICCV2011attr,
BourdevICCV2009, MajiCVPR2011}, which are compound parts consisting of multiple anatomical parts,
highly clustered in 3D configuration space, \eg head and shoulders together. Poselets vote
independently for a hypothesis, and are shown to improve performance. However, they are trained
separately from images annotated specifically in 3D. In contrast, EPM tries to mine out such parts,
at the required atomicity, from given training images for a particular task.
Fig.~\ref{fig:parts_egs} (top right) shows some of the parts for the `female' class which show some
resemblance with poselets, though are not as clean.

Methods such as Poselets and the proposed method are also conceptually comparable to the mid-level
features based algorithms \cite{BoureauCVPR2010, FathiCVPR2008, JooICCV2013, JunejaCVPR2013,
LimCVPR2013, OquabCVPR2014, SabzmeydaniCVPR2007, SinghECCV2012, SunICCV2013}. While Singh \etal
\cite{SinghECCV2012} proposed to discover and exploit mid-level features in a supervised or
semi-supervised way, with alternating between clustering and training discriminative classifiers for
the clusters, Juneja \etal \cite{JunejaCVPR2013} proposed to learn distinctive and recurring image
patches which are discriminative for classifying scene images using a seeding, expansion and
selection based strategy.  Lim \etal \cite{LimCVPR2013} proposed to learn small sketch elements for
contour and object analysis.  Oquab \etal \cite{OquabCVPR2014} used the mid-level features learnt
using CNNs to transfer information to new datasets. Boureau \etal \cite{BoureauCVPR2010} viewed
combinations of popular coding and pooling methods as extracting mid-level features and analysed
them. Sabzmeydani \etal \cite{SabzmeydaniCVPR2007} proposed to learn mid level shapelets features
for pedestrian detection. Yao \etal \cite{YaoICCV2011} proposed to recognize human actions using
bases of human attributes and parts, which can be seen as a kind of mid-level features. The proposed
EPM explores the space of such mid-level features systematically under a discriminative framework
and more distinctively uses only a subset of model parts for scoring \cf all model parts by the
traditional methods. In a recent approach, Parizi \etal \cite{PariziICLR2015} propose to mine out
parts using a $\ell_1/\ell_2$ regularization with weights on parts. They alternate between learning
the discriminative classifier on the pooled part response vector, and the weight vector on the
parts. However, they differ from EPM as they used pooled response of all parts for an image while
EPM considers absolute responses of the best subset of parts from among the collection of an over
complete set of model parts.

Many methods have also been proposed to reconstruct images using patches, \eg Similarity by
Composition by Boiman and Irani \cite{BoimanNIPS2006}, Implicit Shape Models by Leibe \etal
\cite{LeibeIJCV2008}, Naive Bayes Nearest Neighbors (NBNN) by Boiman \etal \cite{BoimanCVPR2008},
and Collaborative Representation by Zhu \etal \cite{ZhuECCV2012}. Similarly sparse representation
has been also used for action recognition in videos \cite{GuhaPAMI2012}. However, while such
approaches are generative and are generally based on minimizing the reconstruction error, EPM aims
to mine out good patches and learn corresponding discriminative templates with the direct aim of
achieving good classification.

\subsection{Description of humans other than actions and attributes} 
Other forms of descriptions of humans have also been reported in the literature. \Eg pose estimation
\cite{AndrilukaCVPR2014, CharlesIJCV2014, DantonePAMI2014, FanCVPR2015, TompsonNIPS2014,
ToshevCVPR2014} and using pose related methods for action \cite{VemulapalliCVPR2014, ThurauCVPR2008,
ChenECCV2012, YaoECCV2012, ZhangCVPR2014} and attribute \cite{ChenECCV2012} recognition have been
studied in computer vision. Recognizing attributes from the faces of humans \cite{BourdevICCV2011,
MaECCVW2012, KumarPAMI2011}, recognizing facial expressions \cite{RudovicPAMI2013, SharmaECCV2012,
WanPR2014, WangICCV2013 } and estimating age from face images \cite{LiCVPR2012, ChangTIP2015,
GengPAMI2007, GuoICCV2009, GuoCVPR2012} have also attracted fair attention. Shao \etal
\cite{ShaoICCV2013} aimed to predict the occupation of humans from images, which can be seen as a
high-level attribute. In the present work, we work with full human bodies where the faces may or may
not be visible and the range of poses may be unconstrained. Although some of the attributes and
actions we consider here are correlated with pose, we do not attempt to solve the challenging
problem of pose first and then infer the said attributes and actions. We directly model such actions
and attributes from the full appearance of the human, expecting the model to make such latent
factorization, implicitly within itself, if required. 

In addition to the works mention above, we also refer the reader to Guo and Lai \cite{GuoPR2014},
for a survey of the general literature for the task of human action recognition from still images.

\section{Expanded Parts Model approach}
\label{sec:approach} 

We address the problem in a supervised classification setting. We assume that a training set of
images and their corresponding binary class labels, \ie 
\begin{align}
\T = \{(\x_i,y_i)| \x_i \in \I, y_i \in \{-1,+1\}, i=1,\ldots, m \}
\end{align} 
are available, where $\I$ is the space of images. We intend to learn a scoring function parametrized
by the model parameters $\Theta$, 
\begin{align}
s_\Theta:\I \rightarrow \R, \ \ \Theta \in \M,
\end{align}
where $\M$ is a class of models (details below), which takes an image and assigns a real valued
score to reflect the membership of the image to the class. In the following we abuse notation and
use $\Theta$ to denote either the parameters of, or the learnt model itself. We define an Expanded
Parts Model (EPM) to be a collection of discriminative templates, each with an associated scale
space location. Images’ scoring, with EPM, is defined as aggregating the scores of the most
discriminative image regions corresponding to a subset of model parts.  The scoring thus (i) uses a
specific subset (different for different images) of model parts and (ii) only scores the
discriminative regions, instead of the whole image. We make these notions formal in the next section
(Sec.~\ref{sec:formulation}).

\subsection{Formulation as regularized loss minimization} 
\label{sec:formulation}
Our model is defined as a collection of discriminative templates with associated 
locations, \ie 
\begin{align} 
\Theta \in \M = \{(\w,\l) | \w \in \R^{Nd}, \l \in [0,1]^{4N}\}
\end{align}
where $N \in \N$ is the number of parts, $d \in \N$ is the dimension of the appearance descriptor,
\begin{align}
\w =[\w_1,\ldots,\w_N], \ \ \w_p \in \R^{d}, \ \ p=1,\ldots,N
\end{align}
is the concatenation of $p=1,\ldots,N$ part templates and 
\begin{align}
\l = [\l_1,\ldots,\l_N] \in [0,1]^{4N}
\end{align}
is the concatenation of their scale-space positions, with each $\l_p$ specifying a bounding box, \ie
\begin{align}
\l_p=[\tilde{x}_1,\tilde{y}_1,\tilde{x}_2,\tilde{y}_2] \in [0,1]^4, \ \ p=1,\dots,N
\end{align} 
where $\tilde{x}$ and $\tilde{y}$ are fractional multiples of width and height respectively.

We propose to learn our model with regularized loss minimization, over the training set $\T$, with the objective 
\begin{equation}
\label{eqn:objective}
L(\Theta; \T) = \frac{\lambda}{2} ||\w||_2^2 + \frac{1}{m} \sum_{i=1}^m \max(0, 1 - y_i s_\Theta(\x_i)),
\end{equation}
with $s_\Theta(\cdot)$ being the scoring function (Sec.\ \ref{sec:scoring_func}). Our objective is
the same as that of linear support vector machines (SVMs) with hinge loss. The only difference is
that we have replaced the linear score function, \ie 
\begin{align}
\tilde{s}_\w(\x) = \w^\top \x,
\end{align} 
with our scoring function. The free parameter $\lambda \in \R$ sets the trade-off between model
regularization and the loss minimization as in the traditional SVM algorithm.

\subsection{Scoring function}\label{sec:scoring_func}
We define the scoring function as
\begin{subequations}
\label{eqn:scoring_func}
\begin{align}
s_\Theta(\x) = & \max_{\a} \frac{1}{\|\a\|_0} \sum_{p=1}^{N} \alpha_p \w_p^\top f(\x,\l_p) \\
& \textrm{s.t. \; \ }  \|\a\|_0 = k \\
& \; \; \; \; \; \; O_v (\a,\l) \leq \beta,
\end{align}
\end{subequations}
where, $\w_p \in \R^d$ is the template of part $p$ and $f(\x,\l_p)$ is the feature extraction
function which calculates the appearance descriptor of the image $\x$, for the patch specified by
$\l_p$, 
\begin{align}
\a = [\alpha_1, \ldots, \alpha_N] \in \{0,1\}^N
\end{align}
are the binary coefficients which specify if a model part is used to score the image or not,
$O_v(\a,\l)$ measures the extent of overlap between the parts selected to score the image. The
$\ell_0$ norm constraint on $\a$ enforces the use of $k$ parts for scoring while the second
constraint encourages coverage in reconstruction by limiting high overlaps. $k \in \N$ and  $\beta
\in \R$ are free parameters of the model. Intuitively what the score function does is that it uses
each model part $\w_p$ to score the corresponding region $\l_p$ in the image $\x$ and then selects
$k$ parts to maximize the average score, while constraining the overlap measure between the parts to
be less than a fixed threshold $\beta$.

Our scoring function is inspired by the methods of (i) image scoring with learnt discriminative
templates, \eg \cite{FelzenszwalbPAMI2010, HussainBMVC2010} and (ii) those of learnt patch
dictionary based image reconstruction \cite{MairalJMLR2010}. We are motivated by these two
principles in the following way. First, by incorporating latent variables, which effectively amount
to a choice of the template(s) that is (are) being used for the current image, the full-scoring
function can be made nonlinear (piecewise linear, to be more precise) while keeping the interaction
with each template as linear. This allows learning of more complex and nonlinear models, especially
in an Expectation Maximization (EM) type algorithm, where algorithms to learn linear templates can
be used once the latent variables are fixed, \eg \cite{FelzenszwalbPAMI2010, HussainBMVC2010}.
Second, similar to the learnt patch dictionary-based reconstruction, we want to have a spatially
distributed representation of the image content, albeit in a discriminative sense, where image
regions are treated independently instead of working with a monolithic global model. With a
discriminative perspective, we would only like to score promising regions, and use only a subset of
model parts, in the images and ignore the background or non-discriminative parts. Exploiting this
could be quite beneficial especially as the discriminative information for human actions and
attributes is often localized in space, \ie for `riding horse' only the rider and the horse are
discriminative and not the background and for `wearing shorts' only the lower part of the (person
centric) image is important. In addition, the model could be over-complete and store information
about the same part at different resolutions, which could lead to possible over-counting, \ie
scoring same image region multiple times with different but related model parts, as well; not
forcing the use of all model parts can help avoid this over-counting.

Hence, we design the scoring function to score the images with the model parts which are most
capable of explaining the possible presence of the class in the image, while (i) using only a subset
of relevant parts from the set of all model parts and (ii) penalizing high overlap of parts used, to
exploit localization and avoid over-counting as discussed above. We aim, thus, to score the image
content only partially (in space) with the most important parts only.

We confirm such behavior of the model with qualitative results in Sec.~\ref{sec:exp_qual}.

\subsection{Solving the optimization problem}
\label{sec:solve}
We propose to solve the model optimization problem using stochastic gradient descent. We use the
stochastic approximation to the sub-gradient \wrt $\w$ given by, 
\begin{align}
\nabla_\w L = \lambda \w -  \delta_i \ \frac{1}{\|\a\|_0} \left[ 
              \begin{array}{c}  
                    \alpha_1 f(\x,\l_1) \\ 
                    \vdots \\ 
                    \alpha_N f(\x,\l_N) 
              \end{array} \right]
\end{align}
where, $\alpha_p$ are obtained by solving Eq.\ \ref{eqn:scoring_func} and
\begin{align}
\delta_i= \begin{cases} 1 &\mbox{if } y_i s_\Theta(\x) < 1 \\ 0 &\mbox{otherwise}. \end{cases}
\end{align}
Alg.~\ref{algo:stoc_grad} gives the pseudo-code for our learning algorithm. The algorithm proceeds
by scoring (and thus calculating the $\a$ for) the current example with $\w$ fixed, and then
updating $\w$ with $\a$ fixed, like in a traditional EM like method.

The scoring function is a constrained binary linear program which is NP-hard. Continuous relaxations
is a popular way of handling such optimizations, \ie relax the $\alpha_i$ to be real in the interval
$[0,1]$ and replace $\| \a\|_0$ with $\| \a \|_1$, and then solve the resulting continuous
constrained linear program and obtain the binary values by thresholding/rounding the continuous
optimum obtained. However, managing the overlap constraint with continuously selected parts would
require additional thought. We instead, decide to take a simpler and direct route via an approximate
greedy approach. Starting with an empty set of selected parts, we greedily add to it the best
scoring part which does not overlap appreciably with all the currently selected parts, for the
current image. The overlap is measured using intersection over union \cite{Everingham2011} and two
parts are considered to overlap significantly with each other if their intersection over union is
more than 1/3.
During training we have an additional constraint on scoring, \ie $\a^\top J \leq \textbf{1}$, where
$J \in \{0,1\}^{N \times m}$ with $J(p,q)=1$ if $p^{th}$ part was sampled from the $q^{th}$ training
image, 0 otherwise. The constraint is enforced by ignoring all the parts that were initialized from
the training images of the currently selected parts. This increases the diversity of learned parts,
by discouraging similar or correlated parts (which emerge from the same training image initially) to
score the current image.  While training, we score each training image from the rest of the train
set, \ie we do not use the model parts which were generated from the same training image to avoid
obvious trivial part selection.

\begin{algorithm}[t]
\centering
\begin{algorithmic}[1]
\STATE \emph{Input}: Training set $\T=\{(\x_i,y_i)\}_{i=1}^m$; denote $m^+$ ($m^-$) as number of positive
(negative) examples
\STATE \emph{Returns}: Learned Expanded Parts Model, $\Theta=(\w,\l)$
\STATE \emph{Initialize}: $\Theta=(\w,\l)$, rate ($\eta_0$), number of parts for scoring ($k$) and
regularization constant ($\lambda$)
\FOR{iter $= 1, \ldots, 10$}
    \STATE $\eta_{+1} \leftarrow \eta_0 \times m^- / m$ and $\eta_{-1} \leftarrow \eta_0 \times m^+ / m$
    \FOR{npass $= 1, \ldots, 5$}
        \STATE $\mathcal{S} \leftarrow$ \emph{rand\_shuffle}($\T$)
        \FORALL{$(\x_i,y_i) \in S$} 
            \STATE Solve Eq.\ \ref{eqn:scoring_func} to get $s_\Theta(\x_i)$ and $\a$
            \STATE $\delta_i \leftarrow $ \emph{binarize}$(y_i s_\Theta(\x_i) < 1)$
            \STATE $\w \leftarrow \w (1 - \eta_{y_i} \lambda) + \delta_i y_i \eta_{y_i} \sum\alpha_p f(\x_i,\l_p) $
        \ENDFOR
    \ENDFOR
    \STATE \texttt{\small parts\_image\_map} $\leftarrow$ \emph{note\_image\_parts} ($\Theta, \mathcal{S}$)
    \STATE $M \leftarrow$ \emph{prune\_parts} ($\Theta$, \texttt{\small parts\_image\_map})
    \STATE \textbf{if} iter $=5$ \textbf{do} \; $\eta \leftarrow \eta/5$ \; \textbf{end if}
\ENDFOR
\caption{SGD for learning Expanded Parts Model (EPM)}
\label{algo:stoc_grad}
\end{algorithmic} 
\end{algorithm} 

Usually, large databases are highly unbalanced, \ie they have many more negative examples than
positive examples (of the order of 50:1). To handle this we use asymmetric learning rates
proportional to the number of examples of other class\footnote{\cite{PerronninCVPR2012} achieve the
same effect by biased sampling from the two classes.} (Step 4, Alg.~\ref{algo:stoc_grad}).

\subsection{Mining discriminative parts}
\begin{figure}[t]
\centering
\includegraphics[width=\columnwidth, trim=0 240 300 0, clip]{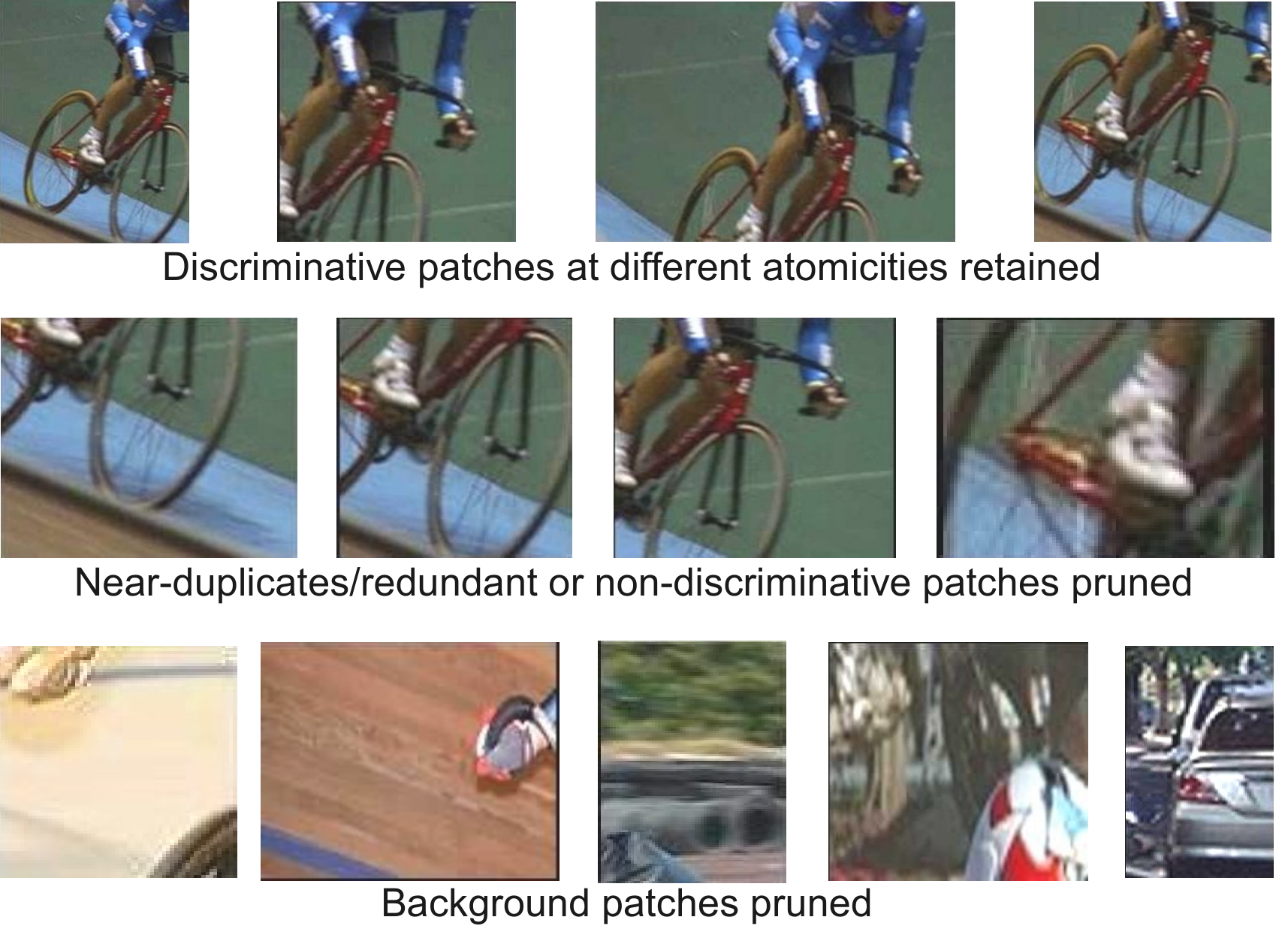} 
\vspace{-1em}
\caption{
Example patches illustrating pruning for the `riding a bike' class. While discriminative patches
(top) at multiple atomicities are retained by the system, redundant or non-discriminative patches
(middle) and random background (bottom) patches are discarded. The patches have been resized and
contrast adjusted, for better visualization. 
}
\label{fig:pruning}
\end{figure}

One of our main intentions is to address important limitations of the current methods: automatically
selecting the task-specific discriminative parts at the appropriate scale space locations. The
search space for finding such parts is very high, as all possible regions in the training images are
potential candidates to be discriminative model parts. We address part mining by two major steps.
First, we resort to randomization for generating the initial pool of candidate model parts. We
randomly sample part candidates from all the training images, to initialize a highly redundant
model. Second, we mine out the discriminative parts from this set by successive pruning. With our
learning set in a stochastic paradigm, we proceed as follows. We first perform a certain number of
passes over randomly shuffled training images and keep track of the parts used while updating them
to learn the model (recall that not all parts are used to score images and, hence, potentially not
all parts in the model, especially when it is highly redundant initially, will be used to score all
the training images). We then note that the parts which are not used by any image will only be
updated due to the regularization term and will finally get very small weights. We accelerate this
shrinking process, and hence the learning process, by pruning them.
Such parts are expected to be either redundant or just non-discriminative background; empirically we
found that to be the case; Fig.~\ref{fig:pruning} shows some examples of the kind of discriminative
parts, at multiple atomicities, that were retained by the model (for `riding a bike' class) while
also some redundant parts as well as background parts which were discarded by the algorithm.

\subsection{Relation with latent SVM}
Our Expanded Parts Model learning formulation is similar to a latent support vector machine (LSVM)
formulation, which optimizes (assuming a hinge loss function) 
\begin{equation}
L(\w; \T) = \frac{\lambda}{2} ||\w||_2^2 + \frac{1}{m} \sum_{i=1}^m \max(0, 1 - y_i s_L(\x_i)),
\end{equation}
where the scoring function is given as 
\begin{align}
s_L(\x) = \max_{\z} \w^\top g(\x,\z),
\end{align}
with $\z$ being the latent variable (\eg part deformations in Deformable Parts-based Model
(DPM)\cite{FelzenszwalbPAMI2010}) and $g(\cdot)$, the feature extraction function. The $\a$, in our
score function Eq.~\ref{eqn:scoring_func}, can be seen as the latent variable (one for each image).
Consequently, the EPM can be seen as a latent SVM similar to the recently proposed model for object
detection by Felzenszwalb \etal \cite{FelzenszwalbPAMI2010}.  

\begin{figure*}
\centering
\hfill
\includegraphics[width=0.315\textwidth, trim=3 0 30 0, clip]{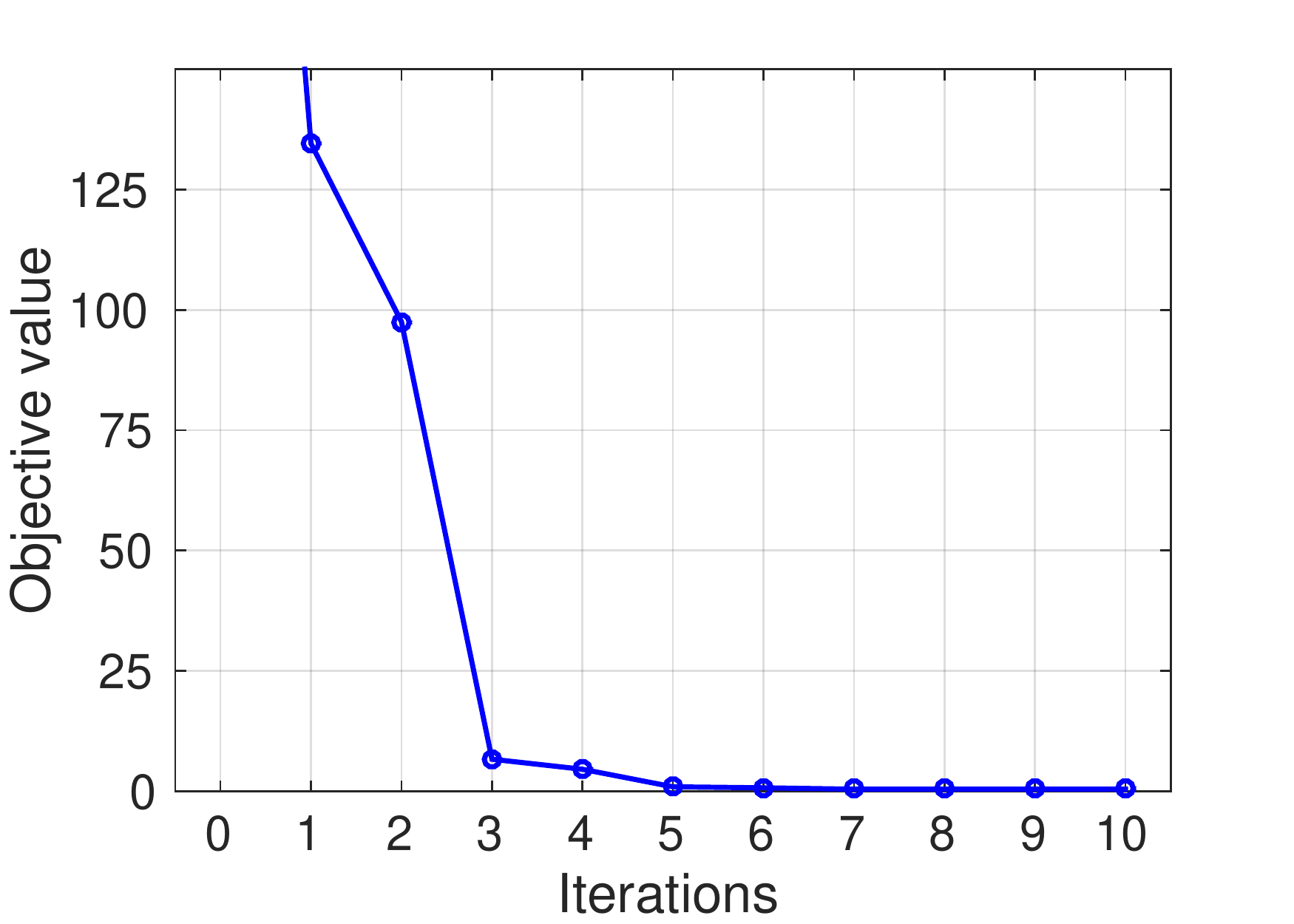} \hfill
\includegraphics[width=0.315\textwidth, trim=5 0 30 0, clip]{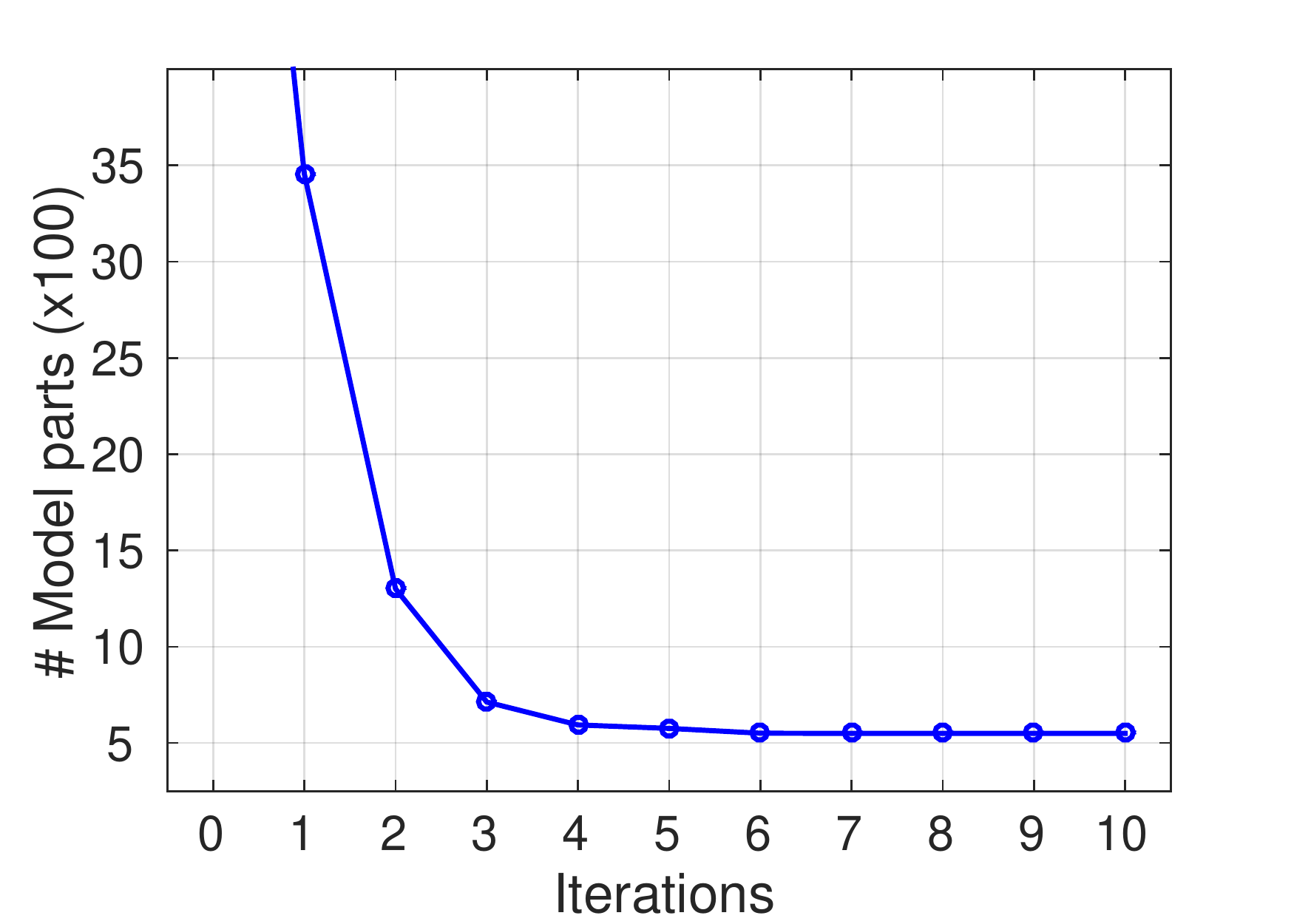} \hfill
\includegraphics[width=0.315\textwidth, trim=5 0 30 0, clip]{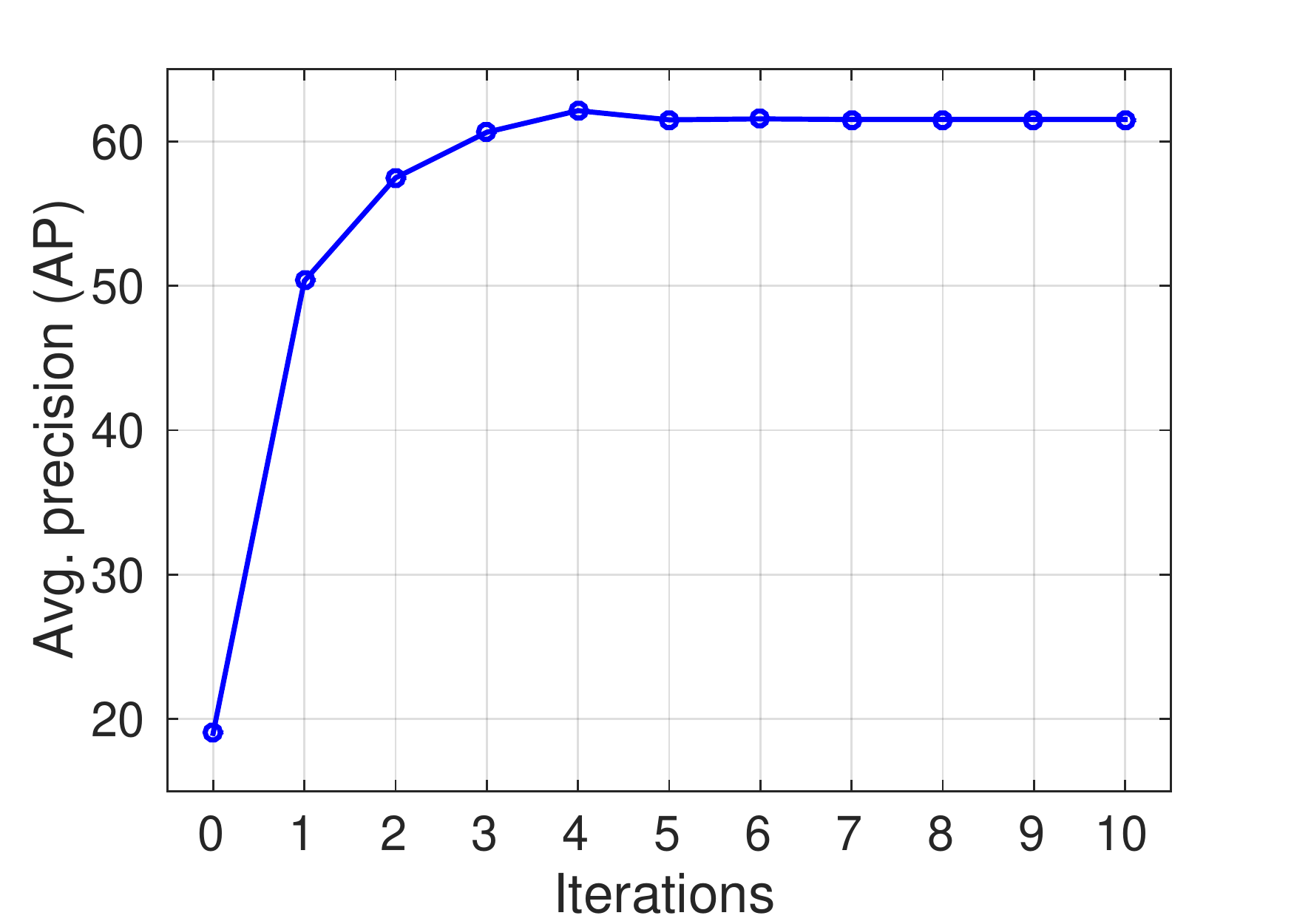} \hfill
\vspace{-1em}
\caption{
The evolution of the (left) objective value, (middle) number of model parts along with the (right)
average precision \vs number of iterations, for the \validation set of `interacting with a computer'
class of the Willow Actions dataset, demonstrating the convergence of our algorithm.  }
\label{fig:objval}
\end{figure*}

In such latent SVM models the objective function is semi-convex \cite{FelzenszwalbPAMI2010}, \ie it
is convex for the negative examples. Such semi-convexity follows from the convexity of scoring
function, with similar arguments as in Felzenszwalb \etal (Sec.\ 4 in \cite{FelzenszwalbPAMI2010}).
The scoring function is a $\max$ over functions which are all linear in $\w$, and hence is convex in
$\w$ which in turn makes the objective function semi-convex. Optimizing while exploiting
semi-convexity gives guarantees that the value of the objective function will either decrease or
stay the same with each update. In the present case, we do not follow Felzenszwalb \etal
\cite{FelzenszwalbPAMI2010} in training, \ie we do not exploit semi-convexity as in practice we did
not observe a significant benefit in doing so. Despite there being no theoretical guarantee of
convergence, we observed that, if the learning rate is not aggressive, training as proposed
leads to good convergence and performance.
Fig.~\ref{fig:objval} shows a typical case demonstrating the convergence of our algorithm, it gives the
value of the objective function, the evolution of the model, in terms of number of parts, and the
performance of the system \vs iterations (Step 4, Alg.~\ref{algo:stoc_grad}), for `interacting with
a computer' class of the Willow Actions dataset.

\subsection{Appearance features and visualization of scoring}

As discussed previously, HOG features are not well adapted to human action recognition. We therefore
resort, in our approach, to using appearance features, \ie the bag-of-features (BoF), for EPM.  When
we use such appearance representation, the so-obtained discriminative models (similar to
\cite{LazebnikCVPR2006}) cannot be called templates (cf.\ HOG based templates \cite{DalalCVPR2005}).
Thus, in the following, we use the word template to loosely denote the similar concept in the
appearance descriptor space.  Note, however, that the proposed method is feature-agnostic and can be
potentially used with any arbitrary appearance descriptor, \eg BoF \cite{CsurkaSLCV2004,
SivicICCV2003}, HOG \cite{DalalCVPR2005}, GIST \cite{OlivaIJCV2001}, CNN \cite{KrizhevskyNIPS2012}
\etc

Since we initialize our parts with the appearance descriptors (like BoF) of patches from training
images (see Sec.~\ref{sec:exp} for details), we can use the initial patches to visualize the scoring
instead of the final learnt templates as in the HOG case. This is clearly a loose association as the
initial patches evolve with training iterations to give the part templates $\w_p$. However we hope
that the appearance of the initial patch will suffice as a proxy for visualizing the part. We found
such an approximate strategy to give reasonable visualizations, \eg Fig.\ \ref{fig:illus_recons}
shows some visualizations of scoring for different classes. While the averaging is not very good,
the visualizations do give an approximate indication of which kind of image regions are scored and
by which kinds of parts. We discuss these more in the qualitative results Sec.~\ref{sec:exp_qual}.

\subsection{Efficient computation using integral histograms}
\label{sec:inthist}
Since we work with a large number of initial model parts, \eg $\Order(10^5)$, the implementation of
how such parts are used to score the images becomes an important algorithmic design aspect. In the
na\"{i}ve approach, the scoring will require computing features for $N$ local regions corresponding
to each of the model part. Since $N$ can be very large for the initial over-complete models, this is
intractable. To circumvent this we use integral histograms \cite{PorikliCVPR2005}, \ie 3D data
structure where we keep integral images corresponding to each dimension of the appearance feature.
The concept was initially introduced by Crow \cite{CrowSIGGRAPH1984} as summed area tables for
texture mapping. It has had a lot of successful applications in computer vision as well
\cite{ViolaIJCV2001, BayCVIU2008, VekslerCVPR2003, AdamCVPR2006}.  

We divide the images with axis aligned regular grid containing rectangular non-overlapping cells.
Denote the location of the lattice points of the grid by $ \mathcal{X}^g = \{ x^g_1, \dots, x^g_s
\}, \mathcal{Y}^g = \{ y^g_1, \dots, y^g_t \}, $ where, $x^g,y^g \in [0,1]$ are fractional multiples
of width and height, respectively. We compute the BoF histograms for image regions from $(0,0)$ to
each of lattice points $(x_i,y_j)$, \ie we compute the feature tensor $F_\x \in \R^{s \times t
\times d}$, for each image $\x$, where the $d$ dimensional vector corresponding to $F(i,j,:)$ is the
corresponding un-normalized BoF vector. When we do random sampling to get candidate parts to
initialize the model (details in Sec.~\ref{sec:exp}), we align the parts to the grid, \ie $ \l_p  =
[\tilde{x}_1,\tilde{y}_1,\tilde{x}_2,\tilde{y}_2], \textrm{s.t. }  \tilde{x}_1  = x^g_i, \
\tilde{y}_1=y^g_j, \tilde{x}_2 = x^g_k, \ \tilde{y}_2=y^g_l,  \forall \textrm{ some }  i,k  \in \{1,
\ldots, s\} \textrm{ and } j,l \in \{1, \ldots, t\} $.

Hence, to score an image with a part we can efficiently compute the feature for the corresponding
location as, 
\begin{align}
f(\x,\l_p) = & F_\x(x^g_k,y^g_l,:) + F_\x(x^g_i,y^g_j,:) \nonumber \\
             & - F_\x(x^g_i,y^g_l,:) - F_\x(x^g_k,y^g_j,:).
\end{align}
$f(\x,\l_p)$ is then normalized appropriately before computing the score by a dot product with
$\w_p$. In this way we do not need to compute the features from scratch, for all regions
corresponding to the model parts every time an image needs to be scored. Also, this way we need to
cache a fixed amount of data, \ie tensor $F_\x$ for every image $\x$.

\section{Experimental results}
\label{sec:exp}
We now present the empirical results of the different experiments we did to validate and analyze the
proposed method. We first give the statistics of the datasets then give implementation details of
our approach as well as our baseline and, finally, proceed to present and discuss our results on the
three datasets.

\vspace{0.8em} \noindent 
\textbf{The datasets.}
We validate and empirically analyze our method on three challenging publicly available datasets:
\vspace{0.3em}

\begin{enumerate}[leftmargin=*]
\item \textit{Willow 7 Human Actions}\footnote{http://www.di.ens.fr/willow/research/stillactions/}
\cite{DelaitreBMVC2010} is a challenging dataset for action classification on unconstrained consumer
images downloaded from the internet. It has 7 classes of common human actions, \eg `ridingbike',
`running'. It has at least 108 images per class of which 70 images are used for training and
validation and the rest are used for testing. The task is to predict the action being performed
given the human bounding box. 
\item \textit{27 Human Attributes (HAT)}\footnote{http://jurie.users.greyc.fr/datasets/hat.html}
\cite{SharmaBMVC2011} is a dataset for learning semantic human attributes. It contains 9344
unconstrained human images obtained by applying a human detector \cite{FelzenszwalbPAMI2010} on
images downloaded from the internet. It has annotations for 27 attributes based on sex, pose (\eg
standing, sitting), age (\eg young, elderly) and appearance (\eg wearing a tee-shirt, shorts). The
dataset has train, validation and test sets. The models are learnt with the \train and \validation
sets and the performance is reported on the \test set. 
\item \textit{Stanford 40 Human
Actions}\footnote{http://vision.stanford.edu/Datasets/40actions.html} \cite{YaoICCV2011} is a
dataset of human actions with 40 diverse daily human actions, \eg brushing teeth, cleaning the
floor, reading books, throwing a frisbee. It has 180 to 300 images per class with a total of 9352
images. We used the suggested \train and \test split provided by the authors on the website, with
100 images per class for training and the rest for testing. 
\end{enumerate}
\vspace{0.3em}
All images are human-centered, \ie the human is assumed to be correctly detected by a previous stage
of the pipeline. On all the three datasets, the performance is evaluated with average precision (AP)
for each class and the mean average precision (mAP) over all classes.

\vspace{0.8em} \noindent 
\textbf{BoF features and baseline.}
Like previous work \cite{DelaitreBMVC2010,SharmaCVPR2012,YaoCVPR2011} we densely sample grayscale
SIFT features at multiple scales. We use a fixed step size of 4 pixels and use square patch sizes
ranging from 8 to 40 pixels. We learn a vocabulary of size 1000 using k-means and assign the SIFT
features to the nearest codebook vector (hard assignment). We use the VLFeat library
\cite{Vedaldi2008} for SIFT and k-means computation.  We use a four-level spatial pyramid with
$\mathcal{C} = \{c \times c | c=1,2,3,4 \}$ cells \cite{LazebnikCVPR2006} as a baseline.  To have
non-linearity we use explicit feature map \cite{VedaldiCVPR2010} with the BoF features.  We use a
map corresponding to the Bhattacharyya kernel, \ie we take dimension-wise square roots of our
$\ell_1$ normalized BoF histograms obtaining $\ell_2$ normalized vectors, which we use with the
baseline as well as with our algorithm. The baseline results are obtained with the liblinear
\cite{FanJMLR2008} library.

\vspace{0.8em} \noindent 
\textbf{Context.}
The immediate context around the person, which might contain partially an associated object (\eg
horse in riding horse) and/or correlated background (\eg grass in running), has been shown to be
beneficial for the task \cite{DelaitreBMVC2010,SharmaCVPR2012}. To include immediate context we
expand the human bounding boxes by 50\% in both width and height.  The context from the full image
has also been shown to be important \cite{DelaitreBMVC2010}. To use it with our method, we add the
scores from a classifier trained on full images to scores from our method. The full image classifier
uses a 4 level SPM with an exponential $\chi^2$ kernel.

\vspace{0.8em} \noindent 
\textbf{Initialization and regularization constant.}
In the initialization we intend to generate a large number of part candidates, which are
subsequently refined by pruning. To achieve this, we randomly sample the positive training images
for patch positions, \ie $\{\l_p\}$ and initialize our model parts as 
\begin{align}
\w_p =  \left[ \begin{array}{c} 2f(\x,\l_p) \\ 1 \end{array}\right],
\ \ p=1,\ldots,N
\end{align}
where $\x$ denotes a BoF histogram. Throughout our method, we append 1 at the end of all our BoF
features to account for the bias term (cf.\ SVM, \eg \cite{PerronninCVPR2012}). This leads to a
score of 1 when a perfect match occurs, 
\begin{align}
\w_p^T \left[ \begin{array}{c}f(\x,\l_p) \\ 1 \end{array}\right] =
[2f(\x,\l_p),1] \left[ \begin{array}{c}f(\x,\l_p) \\ -1 \end{array}\right] = 1,
\end{align} 
and a score of $-1$ in the opposite case, as the appearance features are $\ell_2$-normalized.  For
the learning rate, we follow recent work \cite{PerronninCVPR2012} and fix a learning rate  which we
reduce once for annealing by a factor of 5 halfway through the iterations (Step 15,
Algorithm~\ref{algo:stoc_grad}). We follow \cite{PerronninCVPR2012} and fix the regularization
constant $\lambda=10^{-5}$. 

\vspace{0.8em} \noindent 
\textbf{Deep CNN features.}
Recently, deep Convolutional Neural Networks (CNN) have been very successful, \eg for image
classification \cite{KrizhevskyNIPS2012, SimonyanICLR2015} and object detection
\cite{SzegedyCVPR2015, SermanetCVPR2013, GirshickCVPR2014}, and have been applied for human action
recognition in videos \cite{JiPAMI2013}.  Following such works, we also evaluated the performances
of using the recent highly successful deep Convolutional Neural Networks architectures for image
classification \cite{KrizhevskyNIPS2012, SimonyanICLR2015}. Such networks are trained on large
external image classification datasets such as the Imagenet dataset \cite{DengCVPR2009} and have
been shown to be successful with a large variety of computer vision tasks \cite{RazavianCVPR2014}.
We used the publicly available \texttt{matconvnet} library \cite{Vedaldi2014} and the models,
pre-trained on the Imagenet dataset, corresponding to the network architectures proposed by
Krizhevsky \etal \cite{KrizhevskyNIPS2012} (denoted AlexNet) and by Simonyan and Zisserman
\cite{SimonyanICLR2015} (16 layer network; denoted VGG-16).

\subsection{Quantitative results}
\label{sec:exp_quant}
Tab.\ \ref{tab:wacts_full} shows the results of the proposed Expanded Parts Model (EPM) (with and
without context) along with our implementation of the baseline Spatial Pyramid
\cite{LazebnikCVPR2006} (SPM) and some competing methods using similar features, on the Willow 7
Actions dataset. We achieve a mAP of 66\% which goes up to 67.6\% by adding the full image context.
We perform better than the current state-of-the-art method \cite{SharmaCVPR2012} (with similar
features) on this dataset on five out of seven classes and on average. As demonstrated
by~\cite{DelaitreBMVC2010}, full image context plays an important role in this dataset. It is
interesting to note that even without context, we achieve 3.5\% absolute improvement compared to a
method which models person-object interactions \cite{DelaitreNIPS2011} and uses extra data to train
detectors. 

\begin{table}[t]
\centering
\caption{Performances (mAP) on the Willow Actions dataset 
}
\label{tab:wacts_full}
\begin{tabular}{|r|c|c|c|c|c|c|}
\hline 
Class & \cite{FelzenszwalbPAMI2010} & \cite{DelaitreNIPS2011} & \cite{SharmaCVPR2012} &\cite{LazebnikCVPR2006} & EPM & EPM+C\\
\hline
\hline 
 intr.\ w/ comp.  &  30.2 &  56.6   &    59.7   &     59.7    &   60.8   &    64.5 \\
   photographing  &  28.1 &  37.5   &    42.6   &     42.7    &   40.5   &    40.9 \\
   playing music  &  56.3 &  72.0   &    74.6   &     69.8    &   71.6   &    75.0 \\
     riding bike  &  68.7 &  90.4   &    87.8   &     89.8    &   90.7   &    91.0 \\
    riding horse  &  60.1 &  75.0   &    84.2   &     83.3    &   87.8   &    87.6 \\
         running  &  52.0 &  59.7   &    56.1   &     47.0    &   54.2   &    55.0 \\
         walking  &  56.0 &  57.6   &    56.5   &     53.3    &   56.2   &    59.2 \\
\hline 
\hline
            mean  &  50.2 &  64.1   &    65.9   &     63.7    &   66.0   &    67.6 \\
\hline
\end{tabular}
\end{table}
The second last column in Tab.\ \ref{tab:stan_hat} (upper part) shows our results, with
bag-of-features based representations, along with results of the baseline SPM and other methods, on
the Stanford 40 Actions. EPM performs better than the baseline by 5.8\% (absolute) at 40.7\%.  It
also performs better than Object bank \cite{LiNIPS2010} and Locality-constrained linear coding
\cite{WangCVPR2010llc} (as reported in \cite{YaoICCV2011}) by 8.2\% and 5.5\% respectively. With
context, EPM achieves 42.2\% mAP which is the state-of-the-art result using no external training
data and grayscale features only. Yao \etal \cite{YaoICCV2011} reported higher performance on this
dataset (45.7\%), by performing action recognition using bases of attributes, objects and poses. To
derive their bases they use pre-trained systems for 81 objects, 45 attributes and 150 poselets,
using large amount (comparable to the size of the dataset) of external data. Since they use human
based attributes also, arguably, EPM can be used to improve their generic classifiers and improve
performance further, \ie EPM is complementary to theirs. Khan \etal \cite{KhanIJCV2013} also report
higher (51.9\%) performance on the dataset fusing multiple features, particularly those based on
color, while here we have used only grayscale information. 

The last column in Tab.\ \ref{tab:stan_hat} (upper part) shows ours as well as other results, with
bag-of-features based representations, on the dataset of Human Attributes. Our baseline SPM is
already higher than the results reported by the dataset creators \cite{SharmaBMVC2011}, because we
use denser SIFT and more scales. EPM improves over the baseline by 3.2\% (absolute) and increases
further by 1\% when adding the full image context.  EPM (alone, without context) outperforms the
baseline for 24 out of the 27 attributes. Among the different human attributes, those based on pose
(\eg standing, arms bent, running/walking) are found to be easier than those based on appearance of
clothes (\eg short skirt, bermuda shorts). The range of performance obtained with EPM is quite wide,
from 24\% for crouching to 98\% for standing. 

\begin{figure*}
\centering
\parbox{\textwidth} {
\centering
\begin{tabular}{cc}
\includegraphics[width=.460\linewidth, trim=20 -10 50 25, clip]{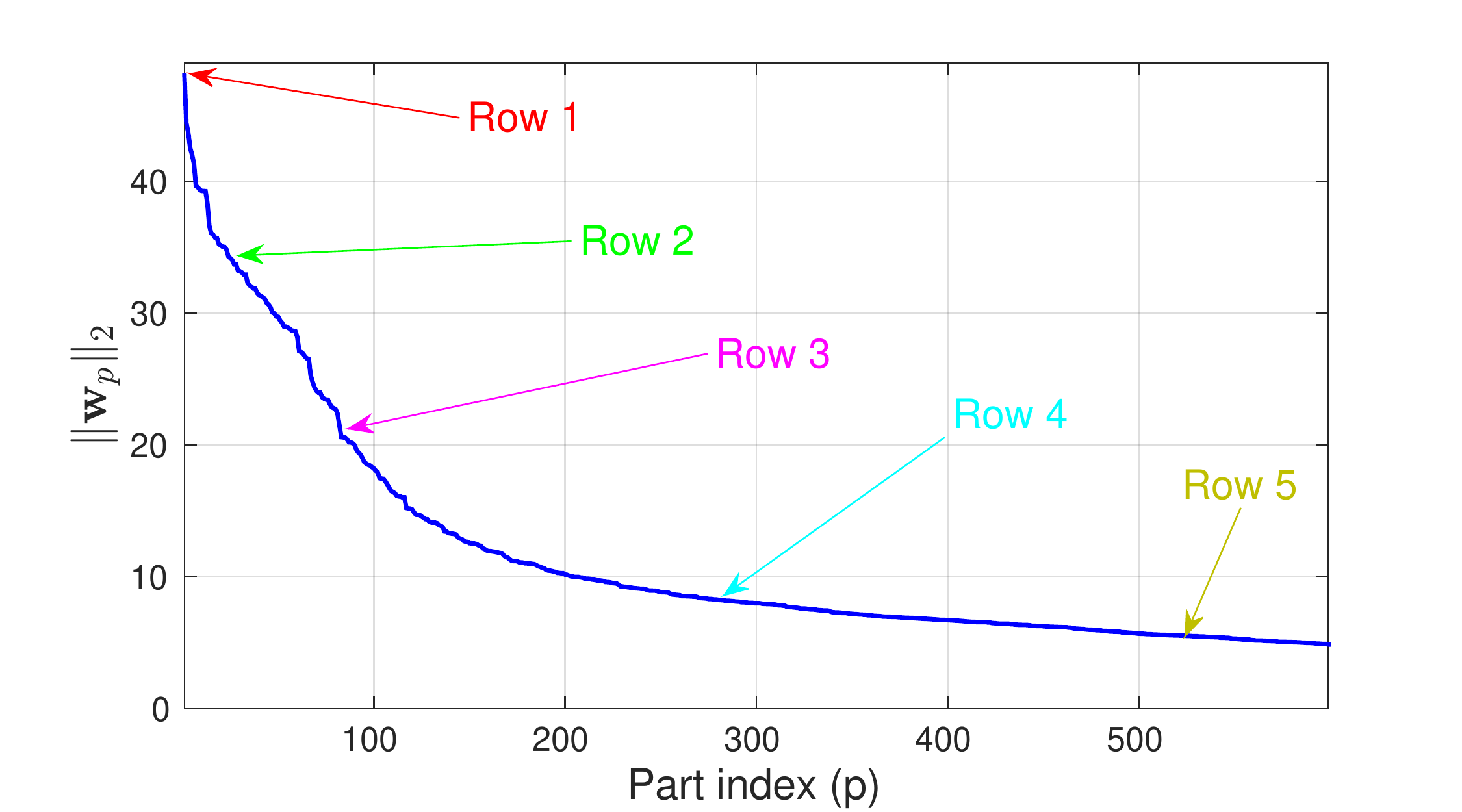} &
\includegraphics[width=.465\linewidth, trim=350 390 122 0, clip]{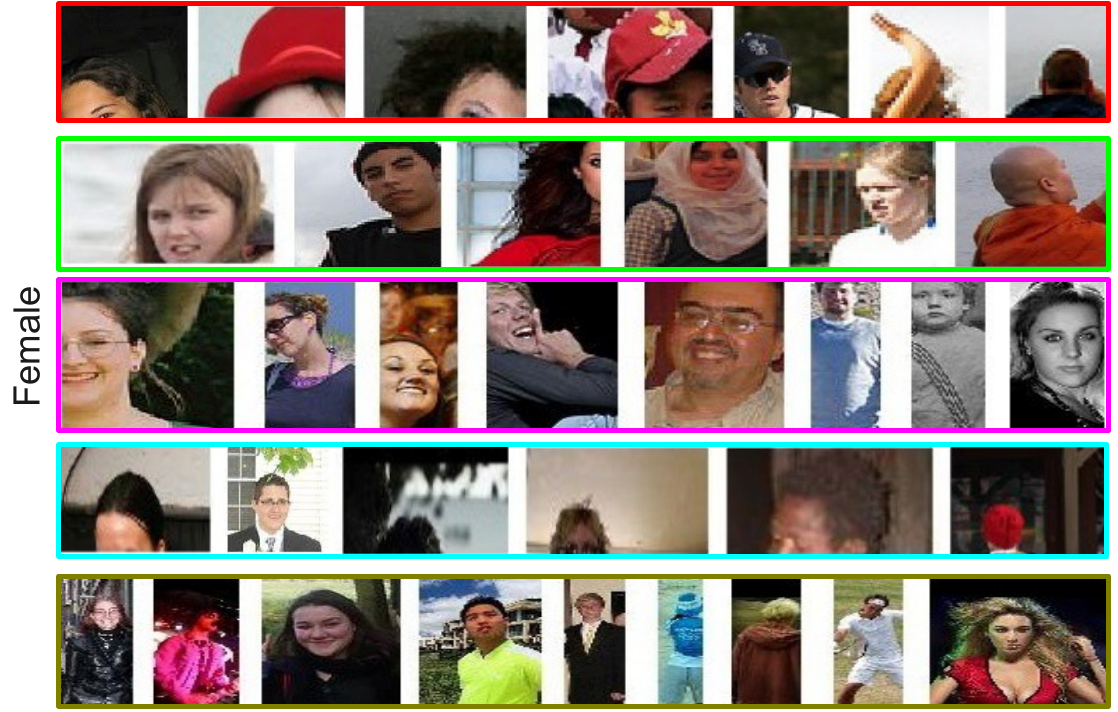} \\
\includegraphics[width=.500\linewidth, trim=0 140 445 220, clip]{parts_egs1} &
\includegraphics[width=.465\linewidth, trim=350 140 122 210, clip]{parts_egs1} \\
\end{tabular}
}
\caption{
Distribution of the norm of the part templates (left top) and some example `parts' (rest three).
Each row illustrates one of the parts: the first image is the patch used to initialize the part and
the remaining images are its top scoring patches.  We show, for each class, parts with different
norms (color coded) of the corresponding $\w_p$ vectors, higher (lower) norm part at top (bottom).
(see Sec.\ \ref{sec:exp_parts} for a discussion, best viewed in color). 
\vspace{-1em}
}
\label{fig:parts_egs}
\end{figure*}

Tab.~\ref{tab:stan_hat} (bottom part) shows the results of the CNN features, on the person bounding
box and the whole image, as well as their combinations with EPM (averaging of the scores of combined
methods), on the two larger datasets, \ie Stanford 40 Actions and Human Attributes.  We can make the
following interesting observations from Tab.~\ref{tab:stan_hat} (bottom part). First, the
performance of the deep features computed on bounding boxes \vs the one on full images follows
inverse trends on the two datasets. On the Stanford Actions dataset, (i) the images are relatively
cleaner (ii) mostly have one prominent person per image and (iii) many classes are scene dependent,
\eg `fixing a bike', `feeding a horse' -- as a result the deep features on the full image perform
better than those on the person bounding boxes. On the other hand, the Human Attributes dataset
often has multiple people per image with more scale variations and the classes are person focused,
\eg `wearing a suit', `elderly', thus, the deep features on the person bounding boxes are better
than those on the full images. Second, we see that the proposed Expanded Parts Model (EPM) based
classifier is not very far in performance from AlexNet (3.3\% and 2\% absolute for the two
datasets). This is quite encouraging as the deep features are trained on large amount of external
data and use the color information of the images, while the EPM is only trained on the respective
training data of the datasets and uses grayscale information only. The stronger VGG-16 network is
much better than both EPM and AlexNet. It is quite interesting to note that EPM is strongly
complementary to the deep features. When using deep features on person bounding boxes, it improves
performance by 6.7\% and 2.7\% on Stanford Actions and by 5.3\% and 4.8\% on Human Attributes, of
AlexNet and VGG-16 networks respectively. When using deep features on full images, the improvements
are 3.4\% and 1.7\% for Stanford Actions and 10.5\% and 8.8\% for Human Attributes datasets.

\begin{table}[t]
\centering
\caption{Performances (mAP) of EPM and deep Convolutional Neural Networks on the Stanford 40 Actions
and the Human Attributes datasets 
}
\label{tab:stan_hat}
\begin{tabular}{|c|c|c|c|}
\hline
Method                         & Image region & Stan40    & \; HAT \;   \\
\hline
\hline
Discr. Spatial Repr.\ \cite{SharmaBMVC2011} & \multirow{3}{*}{bounding box}  &  -   & 53.8 \\
Appearance dict. \cite{JooICCV2013}    & &  -   & 59.3 \\
SPM (baseline) \cite{LazebnikCVPR2006}      & & 34.9 & 55.5 \\
\hline
Object bank \cite{LiNIPS2010}           & full image    & 32.5 & - \\
LLC coding \cite{WangCVPR2010llc}       & bb + full img & 35.2 & - \\
\hline
EPM                                         & bounding box  & 40.7 &  58.7   \\
EPM + Context                               & bb + full img & 42.2 & 59.7 \\
\hline 
\hline
AlexNet (B) \cite{KrizhevskyNIPS2012} & 
\multirow{2}{*}{bounding box}                  &   44.0    &  60.7   \\
VGG-16 (B) \cite{SimonyanICLR2015}    &        &   61.3    &  64.8   \\
\hline
AlexNet (I)                       &  
\multirow{2}{*}{full image}                    &   56.8    &  51.7   \\
VGG-16 (I)                        &            &   70.6    &  55.4   \\
\hline
EPM + AlexNet (B)                 &            
\multirow{2}{*}{bounding box}                  &   50.7    &  66.0   \\
EPM + VGG-16 (B)                  &            &   64.0    &  \textbf{69.6}   \\
\hline
EPM + AlexNet (I)                 &  
\multirow{2}{*}{bb + full img}                  &   60.2    &  62.2   \\
EPM + VGG-16 (I)                  &            &   \textbf{72.3}    &  64.2   \\
\hline 
\end{tabular}
\end{table}

As deep features are not additive like bag-of-features histograms (feature for two image regions
together is not the sum of features for each separately) we can't use the integral histograms based
efficient implementation with the deep features and computing and caching features for all candidate
parts is prohibitive. Hence, we can't use the deep features out-of-the-box with our method.
Tailoring EPM for use with deep architectures is an interesting extension but is out of scope of the
present work.

\subsection{Qualitative results}
\label{sec:exp_qual}
We present qualitative results to illustrate the scoring, Fig.~\ref{fig:illus_recons} shows some
examples, \ie composite images created by averaging the part patches with non-zero alphas. We can
observe that the method focuses on the relevant parts, such as torso and arms for `bent arms',
shorts and tee-shirts for `wearing bermuda shorts', and even computer (left bottom) for `using
computer'. Interestingly, we observe that for both `riding horse' and `riding bike' classes, the
person gets ignored but the hair and helmet have been used partially for scoring. We explain this
with the discriminative nature of the learnt models:  as people in similar pose might confuse the
two classes, the models ignore it and focus on other more discriminative aspects.

\begin{figure*}[t]
\begin{tabular}{ccc}
\centering
\includegraphics[width=0.315\textwidth, trim=10 0 30 0, clip]{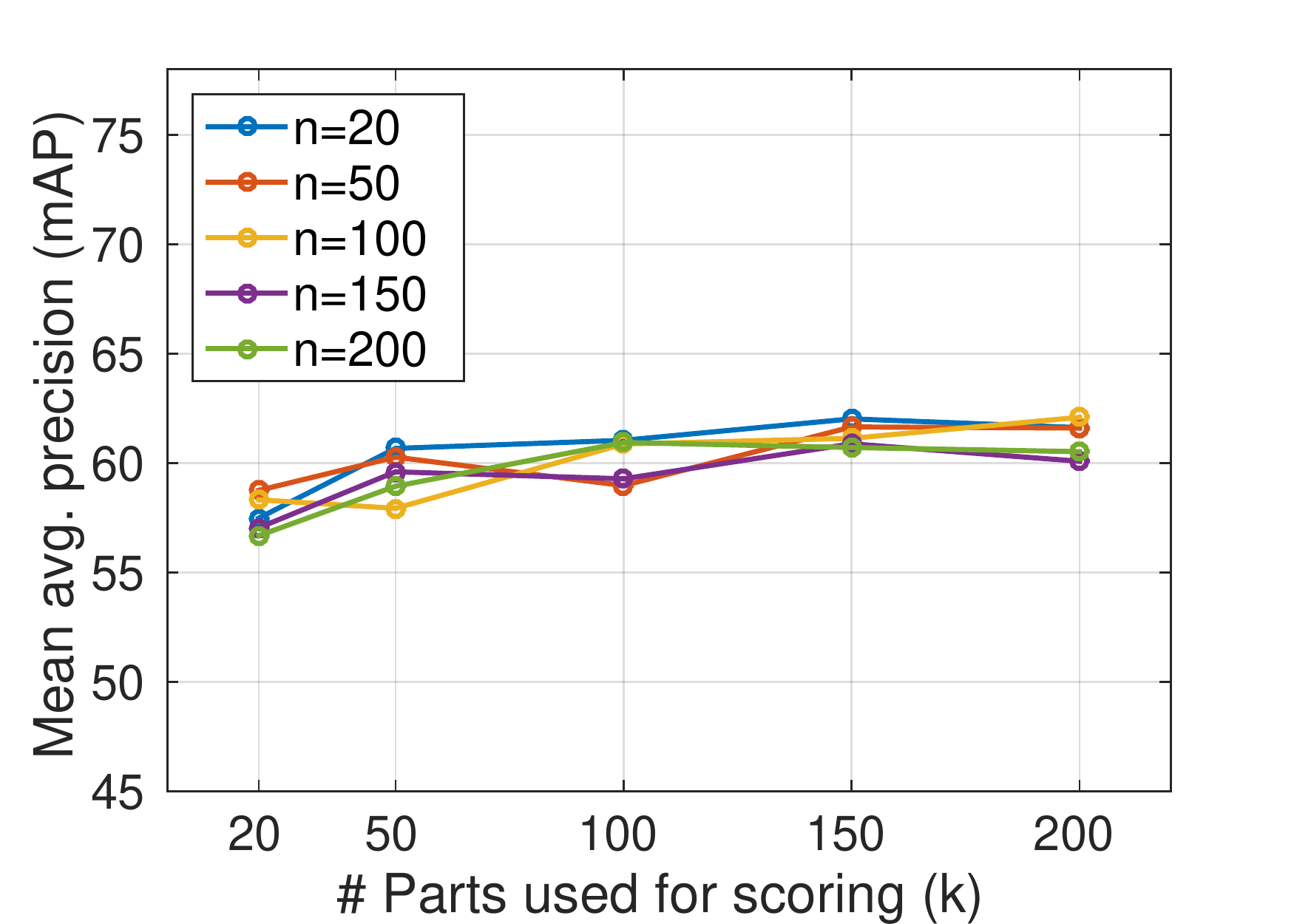} &
\includegraphics[width=0.315\textwidth, trim=10 0 30 0, clip]{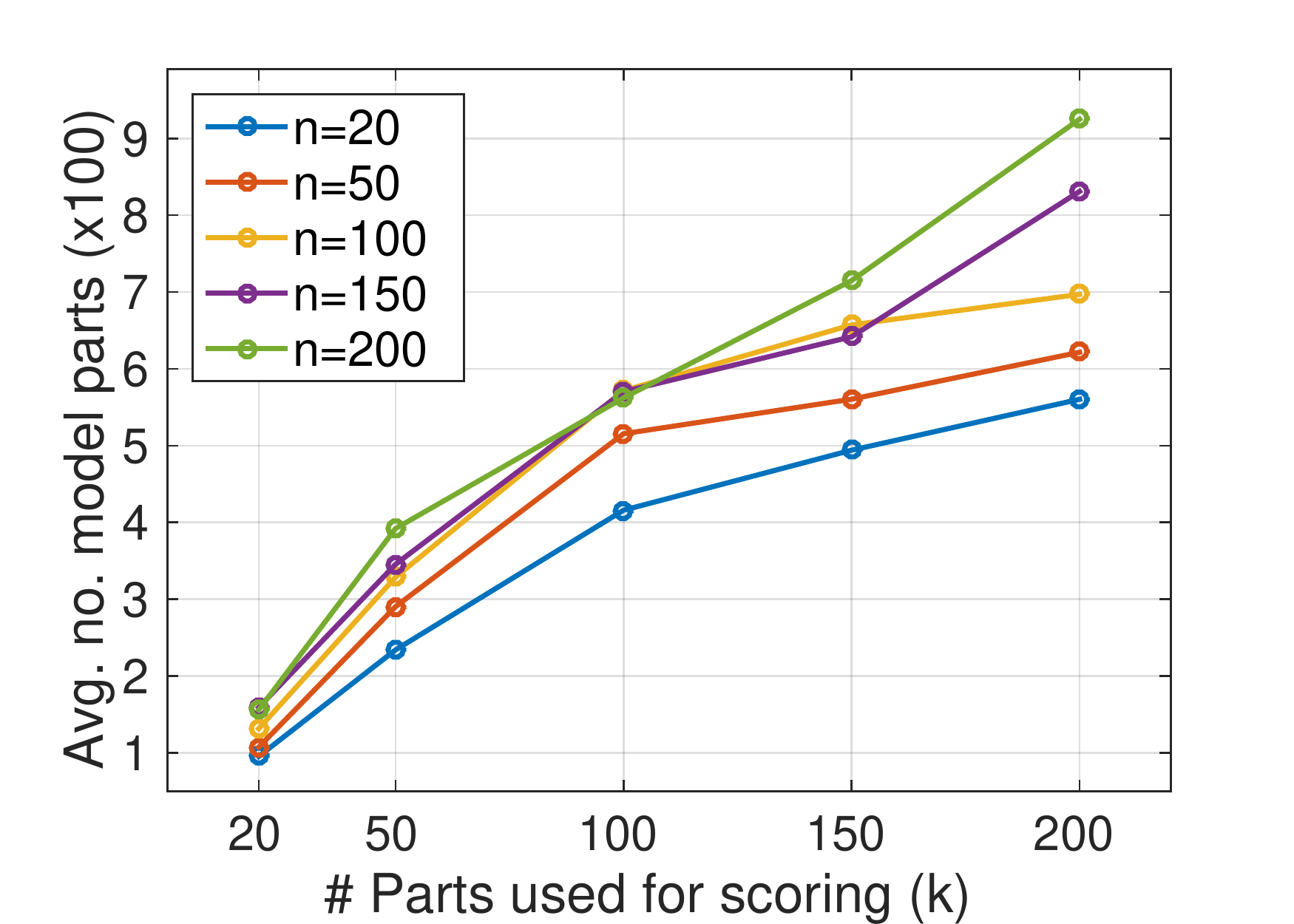}  &
\includegraphics[width=0.315\textwidth, trim=10 0 30 0, clip]{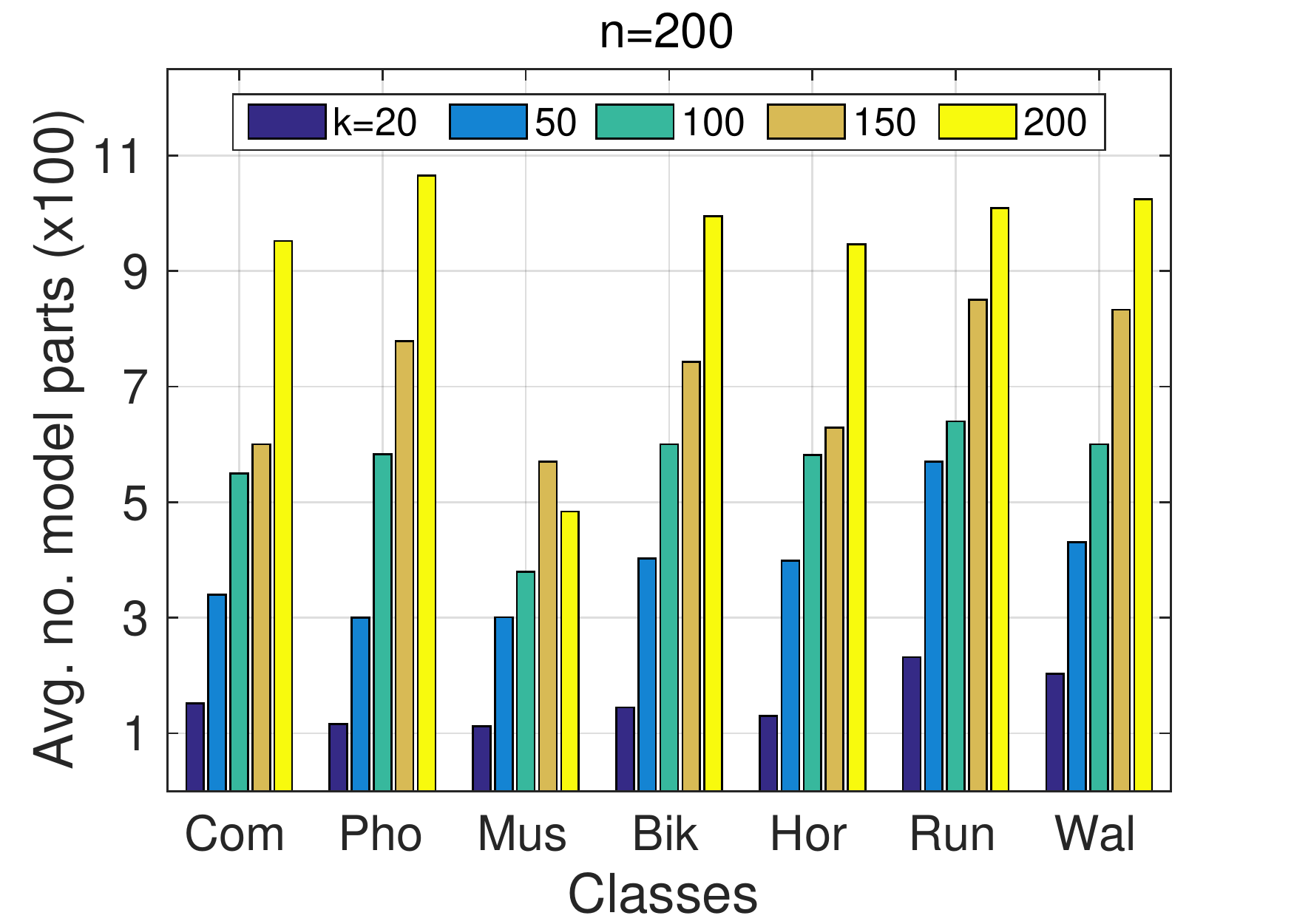} \\
\includegraphics[width=0.315\textwidth, trim=10 0 30 0, clip]{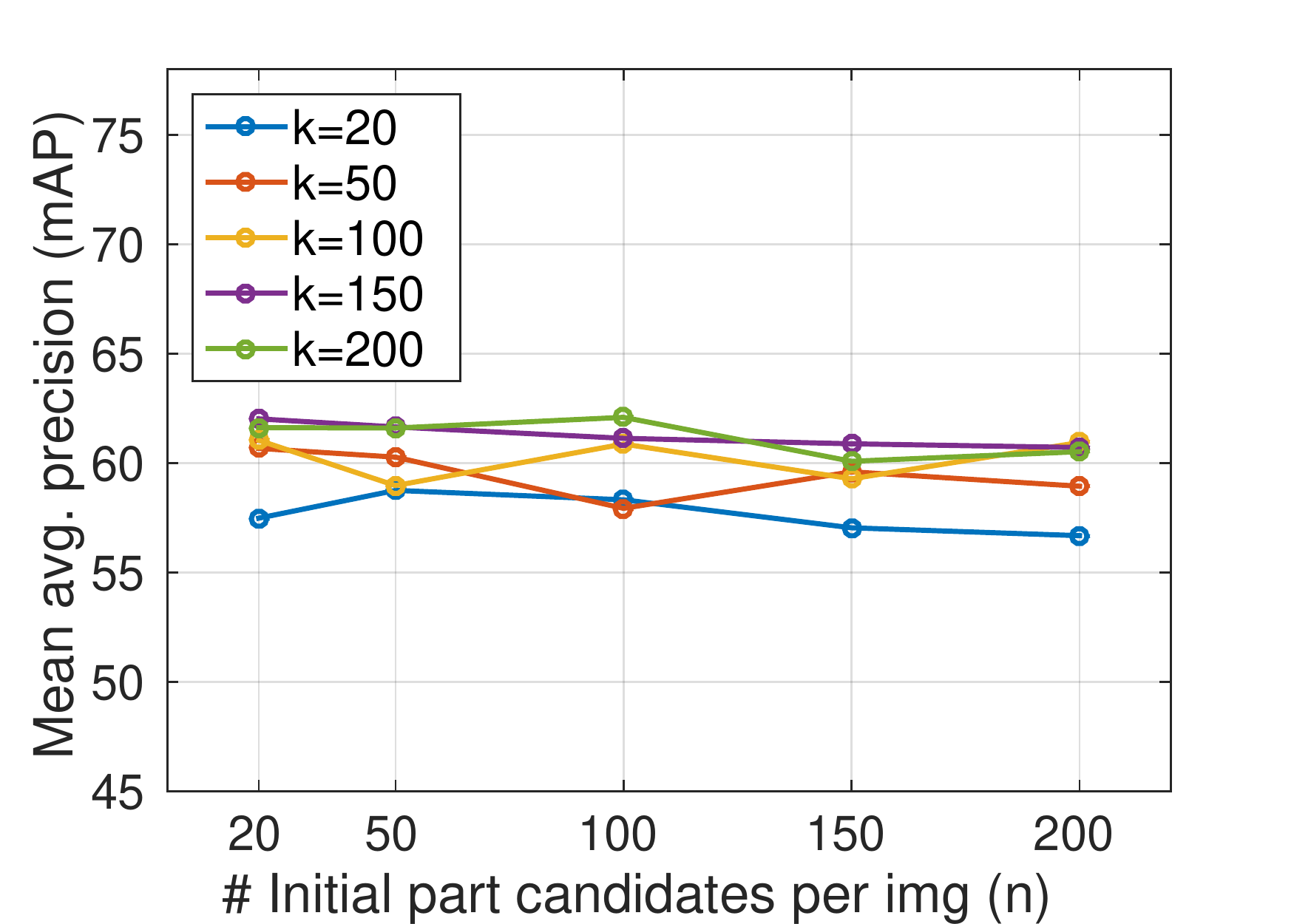} &
\includegraphics[width=0.315\textwidth, trim=10 0 30 0, clip]{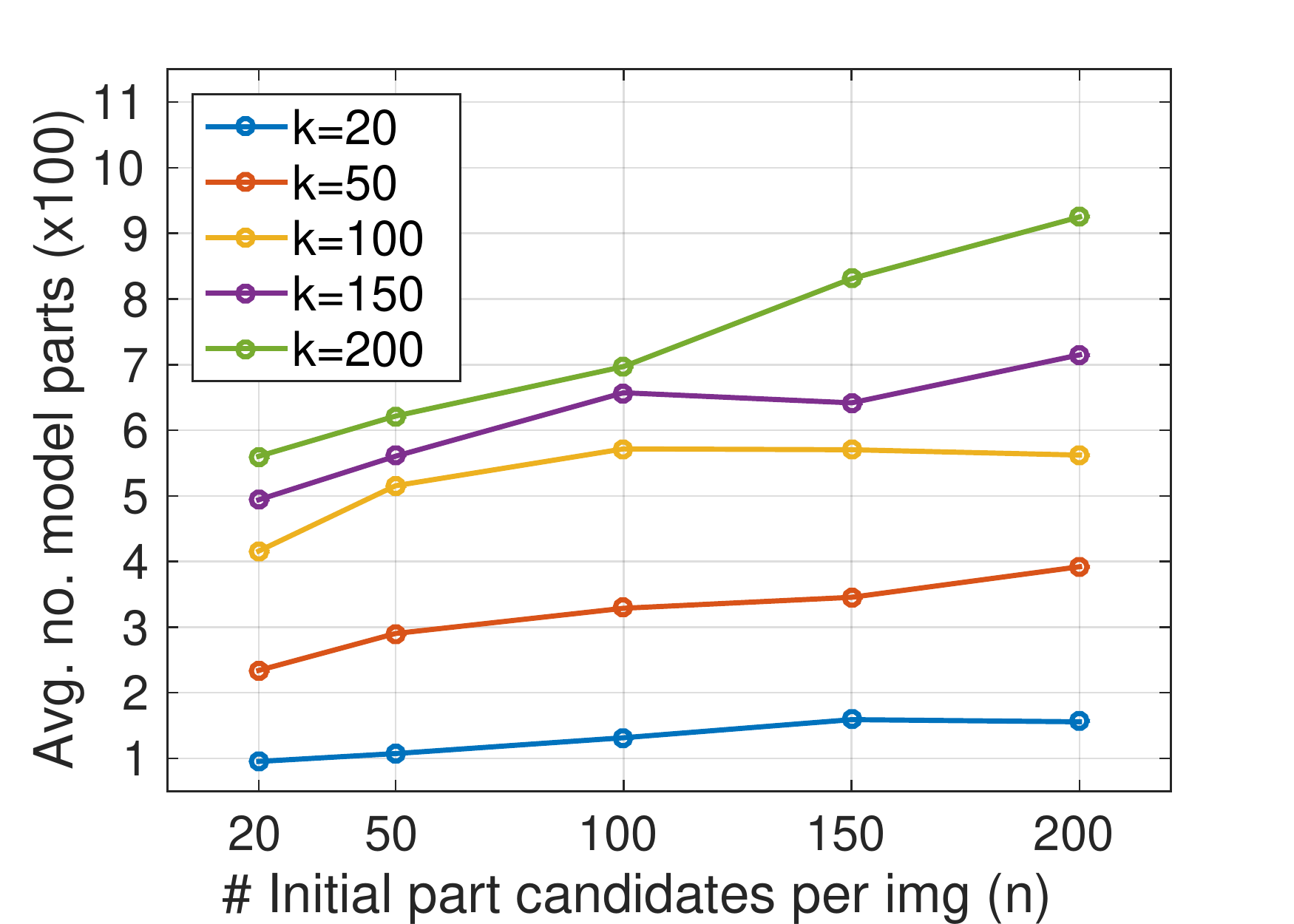} &
\includegraphics[width=0.315\textwidth, trim=10 0 30 0, clip]{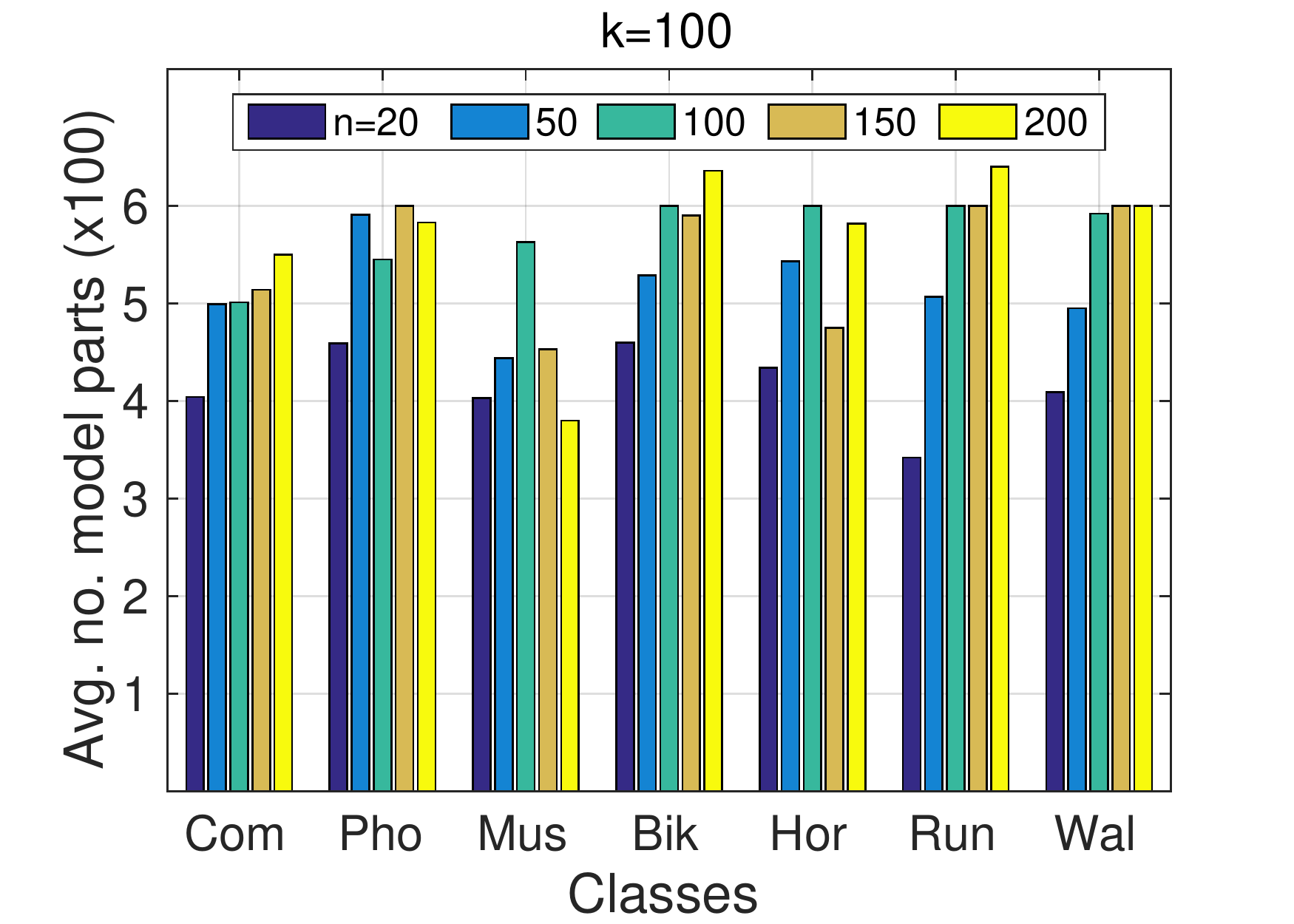} 
\end{tabular}
\caption{
Experiments to evaluate the impact of the number of parts and the number of initial candidate parts
on the performance of the proposed model on the \validation set of the Willow Actions dataset (see
Tab.~\ref{tab:wacts_full} for the full class names). The first row shows
the performances and number of model parts for different values of $k$, \ie the maximum number of
model parts used to score a test image, while the second row shows those for varying $n$, \ie the
number of initial part candidates sampled per training image. 
}
\label{fig:nparts}
\end{figure*}

\begin{figure*}[t]
\centering
\hfill
\includegraphics[width=0.315\textwidth, trim=0 0 33 0, clip]{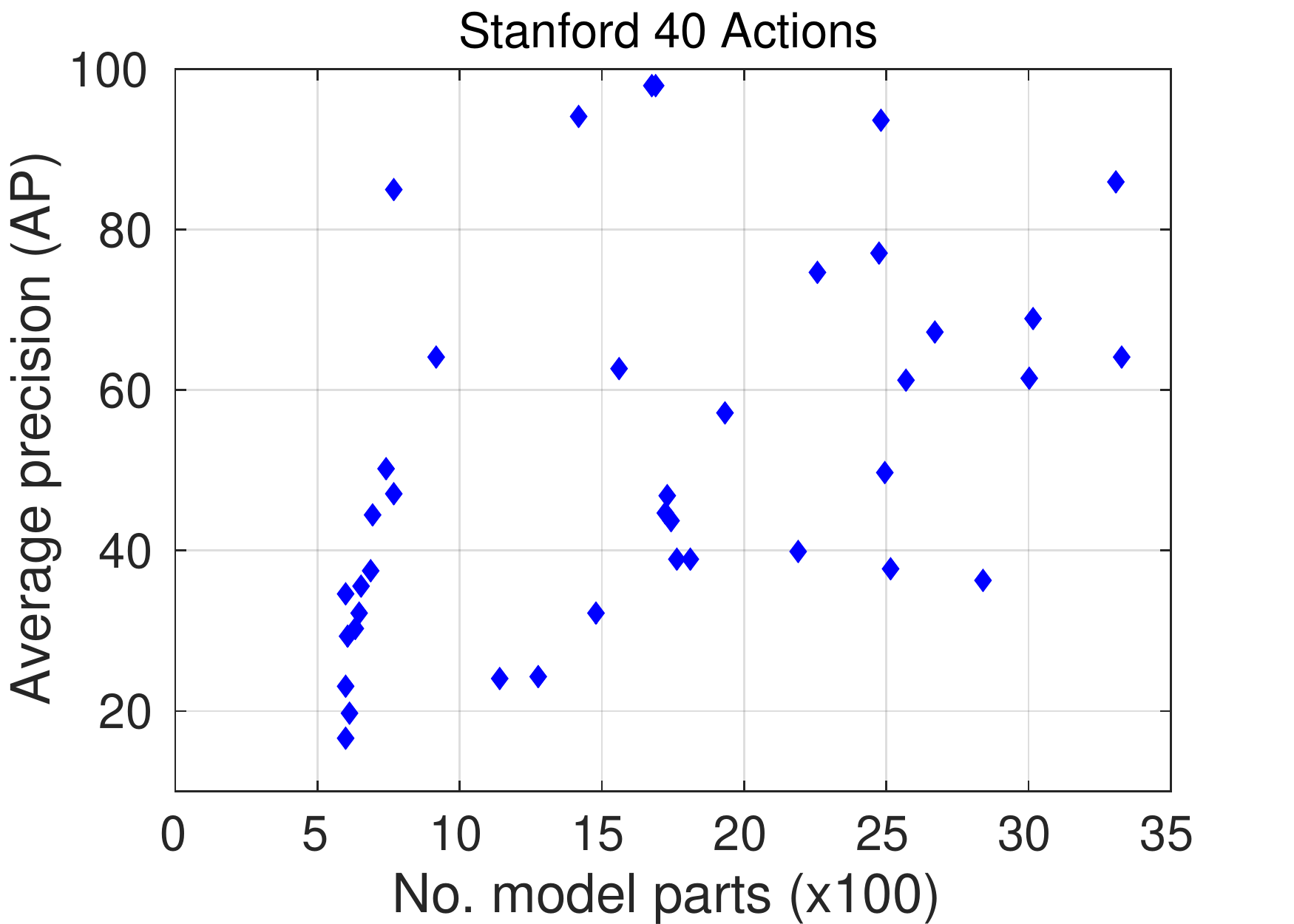} \hfill
\includegraphics[width=0.315\textwidth, trim=0 0 33 0, clip]{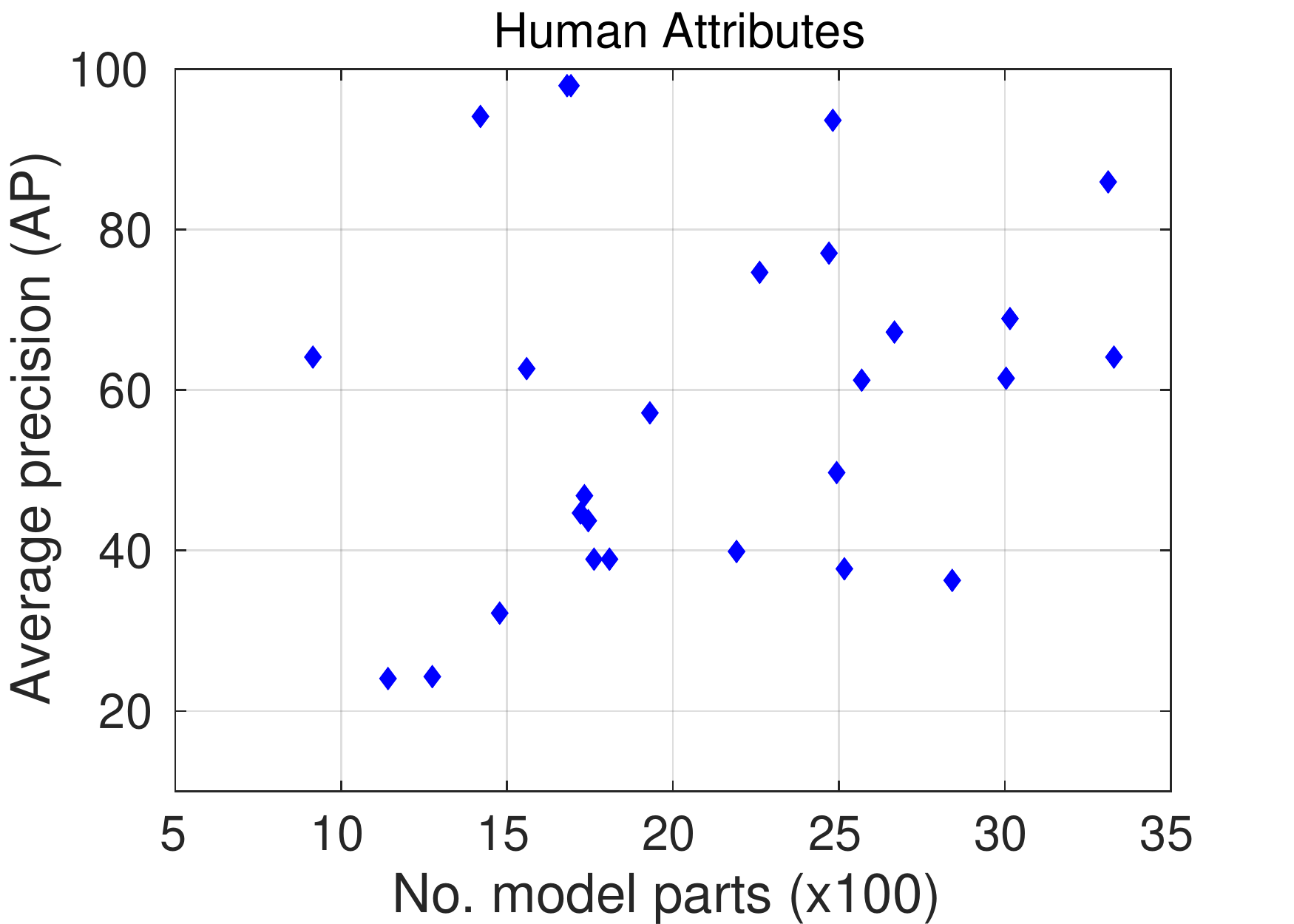} \hfill
\includegraphics[width=0.315\textwidth, trim=0 0 33 0, clip]{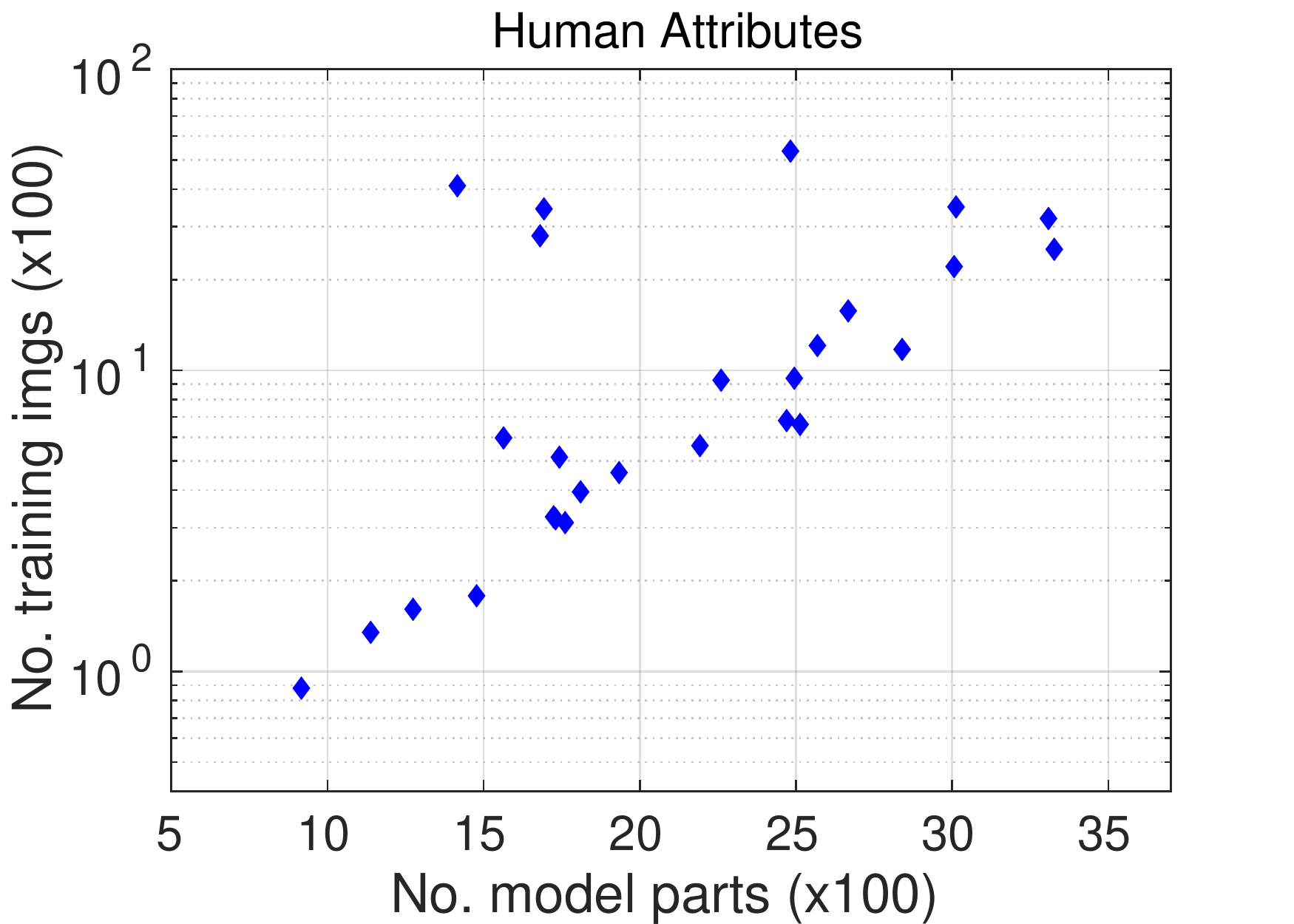} 
\caption{
The average precision obtained by the models for (left) Stanford Actions, (middle) HAT dataset
and (right) the number of training images (for HAT; the number of training images for Stanford
Actions dataset is same for all classes) \vs the number of parts in the final trained models of the
different classes (see Sec.~\ref{sec:exp_parts} for discussion). 
}
\label{fig:np_ap_ntr}
\end{figure*}

\subsection{The parts mined by the model}
\label{sec:exp_parts}
Fig.~\ref{fig:parts_egs} shows the distribution of the $\ell_2$ norm of the learnt part templates,
along with top scoring patches for the selected parts, with norms across the spectrum for three
classes. The first image in any row is the patch with which the part was initialized and the
remaining ones are its top scoring patches. The top scoring patches give an idea of what kind of
appearances the learnt templates $\w_p$ capture. We observe that, across datasets, while most of the
parts seem interpretable, \eg face, head, arms, horse saddle, legs, there are a few parts that seem
to correspond to random background (\eg row 1 for `climbing'). This is in line with a recent study
\cite{ZhuBMVC2012}, in `mixture of template' like formulations, there are clean interpretable
templates along with noisy templates which correspond to background. 

We also observe that the distribution of the $\ell_2$ norm of the parts follows a heavy tailed
distribution. Some parts are very frequent and the system tries to tune them to give high scores for
positive vectors and low scores for negative vectors and hence give them a high overall energy.
There are also parts which have smaller norms, either because they are consistent in appearance
(like the head and partial shoulders on clean backgrounds in row 4 of `female' Fig.\
\ref{fig:parts_egs}, or the leg/arm in the last row of `climbing') or occur in few images. However,
they are discriminative nonetheless.

Fig.~\ref{fig:np_ap_ntr} (left and middle) shows the relation between the performances and the
number of model parts, for the different classes of the larger Stanford Actions and Human Attributes
datasets. The right plot gives the number of training images \vs the number of model parts for the
different classes of the Human Attributes dataset (such curve is not plotted for the Stanford
Actions dataset as it has the same number of training images for each class). We observe that the
model sizes and the performances for the classes are correlated. On the Stanford Actions dataset,
which has the same number of training images for every class, on an average, class models with a
higher number of parts obtain higher performance (correlation coefficient between number of parts
and performances of 0.47). This is somewhat counter intuitive as we would expect that the model with
larger number of parts, and hence larger number of parameters/higher capacity, would over-fit \cf
those with smaller number of parts, for the same amount of training data for both cases. However,
this can be explained as follows. The classes where there are large amounts of variations which are
well captured by the \train set, the model admits larger number of parts to explain the variations
and then successfully generalizes to the \test set.  While for classes where the \train set captures
only a limited amount of variation, the model fits on the \train set with a smaller number of parts
but is then unable to generalize well to the \test set with different variations. An intuitive
feeling of such variations can be had by noting the classes which are relatively well predicted, \eg
`climbing', `riding a horse', `holding an umbrella', \vs those that are not so well predicted, \eg
`texting message', `waving hands', 'drinking' -- while the former classes are expected to have more
visual coherence, the latter are expected to be reatively more visually varied.  

Similar correlation of the number of model parts with performances (Fig.~\ref{fig:np_ap_ntr} middle)
is also observed for Human Attributes dataset (albeit weaker with correlation coefficient 0.23).
Since Human Attributes dataset has different number of images for different classes, it allows us to
make the following interesting observation as well. The performances for Human Attributes dataset
are highly correlated with the number of training images (correlation coefficient 0.79), which is
explained simply as the classes with higher number of images have higher chance performance, and the
classifiers are accordingly better in absolute performance. However, the relationship between the
number of training images and the model parts is close to exponential (correlation coefficient
between the $\log$ of number of training images and the number of model parts 0.65). This is
interesting as it is in line with the heavy tailed nature of visual information -- as the number of
images increase the model expands to capture the visual variability quickly initially but as the
training data increases further the model only expands when it encounters rarer visual information
and hence the growth decreases. The three clear outliers where the increase in training images does
not lead to a increase in model size (after a limit) are `upperbody', 'standing', 'arms bent' ---
these classes are also the best-performing classes; they have relatively high number of training
images but still do not need many model parts as they are limited in their (discriminative) visual
variations.

\subsection{Effect of parameters}
\label{sec:exp_param}
There are two important parameters in the proposed algorithm, first, the number of parts used to
score the images $k$ and, second, the number of candidate parts to be sampled for initializing the
model $n$ (per training image). To investigate the behavior of the method \wrt these two parameters,
we did experiments on the \validation set of the Willow Actions dataset.  Fig.~\ref{fig:nparts}
shows the performances and the model sizes (number of parts in the final models) when varying these
two parameters in the range $\{ 20, 50, 100, 150, 200\}$.  We observe that the average number of
model parts increases rapidly as $k$ is increased (Fig.~\ref{fig:nparts} middle-top). This is
expected to a certain extent, as the pruning of the model parts is dependent on $k$; if $k$ is large
then a larger number of parts are used per image while training, and hence more parts will be used,
on an average, and consequently survive pruning. However, the increase in the model size is not
accompanied by a similarly aggressive increase in the validation performance (Fig.~\ref{fig:nparts}
left-top). The average number of model parts for $k=100$ and $n=200$ is $549$. Similar increase in
the model size but with increase in $n$ is more varied for different values of $k$; for lower value
of say $k=20$ the increase in model size with $n$ is subtle when compared to the same for a higher
value of say $k=200$. However, again such increase in model size doesn't bring increase in
validation performance either. It is also interesting to note the behavior of the models of
different classes when varying $k$ and $n$.  The bar graphs on the right of Fig.~\ref{fig:nparts}
show the number of model parts when $n$ is fixed to $200$ and $k$ is varied (top) and when $k$ is
fixed to $100$ and $n$ is varied. In general, as $k$ was increased the models of almost all the
classes grew in number of parts with $n$ fixed, while when $k$ was fixed and more model parts were
made available, the models first grew and then saturated. The only exception to this was the
`playing music' class where the models practically saturated in both cases, perhaps because of
limited appearance variations. The growing of models with increasing $k$ was followed by a slight
drop in the performance, probably due to over-fitting.

Following these experiments and also for keeping a reasonable computational complexity, $k$ was
fixed to $k=100$ for the experiments reported. This is also comparable to the $85$ cells in the
four-level spatial pyramid representation used as a baseline. Similarly, $n$ was fixed to be $n=200$
for the Willow Actions dataset and $n=20$ for the about $10 \times$ larger Stanford Action and Human
Attributes datasets (recall that $n$ is the number of initial candidates parts sampled per training
image) in the experiments reported. 

\subsection{Training/testing times}
The training is significantly slower compared to a standard SPM/SVM baseline, i.e. by around two
orders of magnitude. This is due to the fact that there is SVM equivalent cost (with a larger number
of vectors) at each iteration.  Testing is also a bit slower compared to an SPM, as it is based on a
dot product between longer vectors. For example, on Stanford dataset testing is 5 times slower
compared to SPM at about 35 milliseconds per image (excluding feature extraction).

\section{Conclusion}
\label{sec:conc}
We have presented a new Expanded Parts Model (EPM) for human analysis. The model learns a collection
of discriminative templates which can appear at specific scale-space positions. It scores a new
image by sparsely explaining only the discriminative regions in the images while using only a subset
of the model parts. We proposed a stochastic sub-gradient based learning method which is efficient
and scalable -- in the largest of our experiments we mine models of $\Order(10^3)$ parts from among
initial candidate sets of $\Order(10^5)$.  We validated our method on three challenging publicly
available datasets for human attributes and actions. We also showed complementary nature of the
proposed method to the current state-of-the-art deep Convolutional Neural Networks based features.
Apart from obtaining good quantitative results, we analysed the nature of the parts obtained and
also analysed the growth of the model size with the complexity of the visual task as well as the
amount of training data available.

\section*{Acknowldgements}
This work was partly realized as part of the Quaero Programme, funded by OSEO, French State agency
for innovation, by the ANR (grant reference ANR-2010-CORD-103-06) and by the ERC advanced grant
ALLEGRO.


%





\ifCLASSOPTIONcaptionsoff
  \newpage
\fi



\bibliographystyle{IEEEtran}
%
\bibliography{biblio}

\begin{thebibliography}{100}
\providecommand{\url}[1]{#1}
\csname url@samestyle\endcsname
\providecommand{\newblock}{\relax}
\providecommand{\bibinfo}[2]{#2}
\providecommand{\BIBentrySTDinterwordspacing}{\spaceskip=0pt\relax}
\providecommand{\BIBentryALTinterwordstretchfactor}{4}
\providecommand{\BIBentryALTinterwordspacing}{\spaceskip=\fontdimen2\font plus
\BIBentryALTinterwordstretchfactor\fontdimen3\font minus
  \fontdimen4\font\relax}
\providecommand{\BIBforeignlanguage}[2]{{%
\expandafter\ifx\csname l@#1\endcsname\relax
\typeout{** WARNING: IEEEtran.bst: No hyphenation pattern has been}%
\typeout{** loaded for the language `#1'. Using the pattern for}%
\typeout{** the default language instead.}%
\else
\language=\csname l@#1\endcsname
\fi
#2}}
\providecommand{\BIBdecl}{\relax}
\BIBdecl

\bibitem{YangCVPR2010}
W.~Yang, Y.~Wang, and G.~Mori, ``Recognizing human actions from still images
  with latent poses,'' in \emph{CVPR}, 2010.

\bibitem{YaoCVPR2010a}
B.~Yao and L.~Fei-Fei, ``Modeling mutual context of object and human pose in
  human-object interaction activities,'' in \emph{CVPR}, 2010.

\bibitem{DelaitreBMVC2010}
V.~Delaitre, I.~Laptev, and J.~Sivic, ``Recognizing human actions in still
  images: {A} study of bag-of-features and part-based representations,'' in
  \emph{BMVC}, 2010.

\bibitem{SharmaBMVC2011}
G.~Sharma and F.~Jurie, ``Learning discriminative representation for image
  classification,'' in \emph{BMVC}, 2011.

\bibitem{SharmaCVPR2012}
G.~Sharma, F.~Jurie, and C.~Schmid, ``Discriminative spatial saliency for image
  classification,'' in \emph{CVPR}, 2012.

\bibitem{YaoCVPR2011}
B.~Yao, A.~Khosla, and L.~Fei-Fei, ``Combining randomization and discrimination
  for fine-grained image categorization,'' in \emph{CVPR}, 2011.

\bibitem{Everingham2011}
M.~Everingham, L.~Van~Gool, C.~K.~I. Williams, J.~Winn, and A.~Zisserman, ``The
  {PASCAL} {V}isual {O}bject {C}lasses {C}hallenge 2011 {(VOC2011)}
  {R}esults,''
  http://www.pascal-network.org/challenges/VOC/voc2011/workshop/index.html,
  2011.

\bibitem{DelaitreNIPS2011}
V.~Delaitre, J.~Sivic, and I.~Laptev, ``Learning person-object interactions for
  action recognition in still images,'' in \emph{NIPS}, 2011.

\bibitem{DesaiCVPRW2010}
C.~Desai, D.~Ramanan, and C.~Fowlkes, ``Discriminative models for static
  human-object interactions,'' in \emph{CVPR Workshops}, 2010.

\bibitem{GuptaPAMI2009}
A.~Gupta, A.~Kembhavi, and L.~S. Davis, ``Observing human-object interactions:
  {U}sing spatial and functional compatibility for recognition,'' \emph{PAMI},
  vol.~31, pp. 1775--1789, October 2009.

\bibitem{PrestPAMI2011}
A.~Prest, C.~Schmid, and V.~Ferrari, ``Weakly supervised learning of
  interactions between humans and objects,'' \emph{PAMI}, vol.~34, no.~3, pp.
  601--614, 2011.

\bibitem{YaoCVPR2010}
B.~Yao and L.~Fei-Fei, ``Grouplet: {A} structured image representation for
  recognizing human and object interactions,'' in \emph{CVPR}, 2010.

\bibitem{OlshausenVR1997}
B.~A. Olshausen and D.~J. Field, ``Sparse coding with an overcomplete basis
  set: A strategy employed by v1?'' \emph{Vision research}, vol.~37, no.~23,
  pp. 3311--3325, 1997.

\bibitem{YangCVPR2009}
J.~Yang, K.~Yu, Y.~Gong, and T.~Huang, ``Linear spatial pyramid matching using
  sparse coding for image classification,'' in \emph{CVPR}, 2009.

\bibitem{YangECCV2010}
J.~Yang, K.~Yu, and T.~Huang, ``Efficient highly over-complete sparse coding
  using a mixture model,'' in \emph{ECCV}, 2010.

\bibitem{MairalJMLR2010}
J.~Mairal, F.~Bach, J.~Ponce, and G.~Sapiro, ``Online learning for matrix
  factorization and sparse coding,'' \emph{JMLR}, vol.~11, pp. 19--60, 2010.

\bibitem{YangTIP2010}
J.~Yang, J.~Wright, T.~S. Huang, and Y.~Ma, ``Image super-resolution via sparse
  representation,'' \emph{TIP}, vol.~19, no.~11, pp. 2861--2873, 2010.

\bibitem{WrightPAMI2009}
J.~Wright, A.~Y. Yang, A.~Ganesh, S.~S. Sastry, and Y.~Ma, ``Robust face
  recognition via sparse representation,'' \emph{PAMI}, vol.~31, no.~2, pp.
  210--227, 2009.

\bibitem{JiaICCV2011}
K.~Jia, X.~Wang, and X.~Tang, ``Optical flow estimation using learned sparse
  model,'' in \emph{ICCV}, 2011.

\bibitem{SharmaCVPR2013}
G.~Sharma, F.~Jurie, and C.~Schmid, ``Expanded parts model for human attribute
  and action recognition in still images,'' in \emph{CVPR}, 2013.

\bibitem{LazebnikCVPR2006}
S.~Lazebnik, C.~Schmid, and J.~Ponce, ``Beyond bags of features: {S}patial
  pyramid matching for recognizing natural scene categories,'' in \emph{CVPR},
  2006.

\bibitem{CsurkaSLCV2004}
G.~Csurka, C.~R. Dance, L.~Fan, J.~Willamowski, and C.~Bray, ``Visual
  categorization with bags of keypoints,'' in \emph{Intl. Workshop on Stat.
  Learning in Comp. Vision}, 2004.

\bibitem{SivicICCV2003}
J.~Sivic and A.~Zisserman, ``{Video Google}: {A} text retrieval approach to
  object matching in videos,'' in \emph{ICCV}, 2003.

\bibitem{LoweIJCV2004}
D.~Lowe, ``Distinctive image features form scale-invariant keypoints,''
  \emph{IJCV}, vol.~60, no.~2, pp. 91--110, 2004.

\bibitem{DalalCVPR2005}
N.~Dalal and B.~Triggs, ``Histograms of oriented gradients for human
  detection,'' in \emph{CVPR}, 2005.

\bibitem{BenensonCVPR2012}
R.~Benenson, M.~Mathias, R.~Timofte, and L.~Van~Gool, ``Pedestrian detection at
  100 frames per second,'' in \emph{CVPR}, 2012.

\bibitem{DollarPAMI2014}
P.~Doll{\'a}r, R.~Appel, S.~Belongie, and P.~Perona, ``Fast feature pyramids
  for object detection,'' \emph{PAMI}, vol.~36, no.~8, pp. 1532--1545, 2014.

\bibitem{FelzenszwalbPAMI2010}
P.~Felzenszwalb, R.~Girshick, D.~McAllester, and D.~Ramanan, ``Object detection
  with discriminatively trained part based models,'' \emph{PAMI}, vol.~32,
  no.~9, pp. 1627--1645, 2010.

\bibitem{PandeyICCV2011}
M.~Pandey and S.~Lazebnik, ``Scene recognition and weakly supervised object
  localization with deformable part-based models,'' in \emph{ICCV}, 2011.

\bibitem{KhanIJCV2013}
F.~S. Khan, R.~M. Anwer, J.~van~de Weijer, A.~D. Bagdanov, A.~M. Lopez, and
  M.~Felsberg, ``Coloring action recognition in still images,'' \emph{IJCV},
  vol. 105, no.~3, pp. 205--221, 2013.

\bibitem{MalisiewiczICCV2011}
T.~Malisiewicz, A.~Gupta, and A.~Efros, ``Ensemble of {Exemplar-SVM}s for
  object detection and beyond,'' in \emph{ICCV}, 2011.

\bibitem{YanECCV2012}
S.~Yan, X.~Xu, D.~Xu, S.~Lin, and X.~Li, ``Beyond spatial pyramids: A new
  feature extraction framework with dense spatial sampling for image
  classification,'' in \emph{ECCV}, 2012.

\bibitem{JainCVPR2013}
M.~Jain, H.~Jegou, and P.~Bouthemy, ``Better exploiting motion for better
  action recognition,'' in \emph{CVPR}, 2013.

\bibitem{JainCVPR2014}
M.~Jain, J.~van Gemert, H.~Jegou, P.~Bouthemy, and C.~G. Snoek, ``Action
  localization with tubelets from motion,'' in \emph{CVPR}, 2014.

\bibitem{OneataCVPR2014}
D.~Oneata, J.~Verbeek, and C.~Schmid, ``Efficient action localization with
  approximately normalized fisher vectors,'' in \emph{CVPR}, 2014.

\bibitem{WangICCV2013}
H.~Wang and C.~Schmid, ``Action recognition with improved trajectories,'' in
  \emph{ICCV}, 2013.

\bibitem{LaptevCVPR2008}
I.~Laptev, M.~Marszalek, C.~Schmid, and B.~Rozenfeld, ``Learning realistic
  human actions from movies,'' in \emph{CVPR}, 2008.

\bibitem{SimonyanNIPS2014}
K.~Simonyan and A.~Zisserman, ``Two-stream convolutional networks for action
  recognition in videos,'' in \emph{NIPS}, 2014.

\bibitem{FergusIJCV2007}
R.~Fergus, P.~Perona, and A.~Zisserman, ``Weakly supervised scale-invariant
  learning of models for visual recognition,'' \emph{IJCV}, vol.~71, no.~3, pp.
  273--303, March 2007.

\bibitem{DesaiECCV2012}
C.~Desai and D.~Ramanan, ``Detecting actions, poses, and objects with
  relational phraselets,'' in \emph{ECCV}, 2012.

\bibitem{YangCVPR2011}
Y.~Yang and D.~Ramanan, ``Articulated pose estimation with flexible
  mixtures-of-parts,'' in \emph{CVPR}, 2011.

\bibitem{ZhuCVPR2012}
X.~Zhu and D.~Ramanan, ``Face detection, pose estimation, and landmark
  localization in the wild,'' in \emph{CVPR}, 2012.

\bibitem{ZhuBMVC2012}
X.~Zhu, C.~Vondrick, D.~Ramanan, and C.~Fowlkes, ``Do we need more training
  data or better models for object detection?'' in \emph{BMVC}, 2012.

\bibitem{BourdevICCV2011}
L.~Bourdev, S.~Maji, and J.~Malik, ``Describing people: Poselet-based attribute
  classification,'' in \emph{ICCV}, 2011.

\bibitem{BourdevICCV2011attr}
------, ``Describing people: A poselet-based approach to attribute
  classification,'' in \emph{ICCV}, 2011.

\bibitem{BourdevICCV2009}
L.~Bourdev and J.~Malik, ``Poselets: Body part detectors trained using {3D}
  human pose annotations,'' in \emph{ICCV}, 2009.

\bibitem{MajiCVPR2011}
S.~Maji, L.~Bourdev, and J.~Malik, ``Action recognition from a distributed
  representation of pose and appearance,'' in \emph{CVPR}, 2011.

\bibitem{BoureauCVPR2010}
Y.-L. Boureau, F.~Bach, Y.~LeCun, and J.~Ponce, ``Learning mid-level features
  for recognition,'' in \emph{CVPR}, 2010.

\bibitem{FathiCVPR2008}
A.~Fathi and G.~Mori, ``Action recognition by learning mid-level motion
  features,'' in \emph{CVPR}, 2008.

\bibitem{JooICCV2013}
J.~Joo, S.~Wang, and S.-C. Zhu, ``Human attribute recognition by rich
  appearance dictionary,'' in \emph{ICCV}, 2013.

\bibitem{JunejaCVPR2013}
M.~Juneja, A.~Vedaldi, C.~Jawahar, and A.~Zisserman, ``Blocks that shout:
  Distinctive parts for scene classification,'' in \emph{CVPR}, 2013.

\bibitem{LimCVPR2013}
J.~J. Lim, C.~L. Zitnick, and P.~Doll{\'a}r, ``Sketch tokens: A learned
  mid-level representation for contour and object detection,'' in \emph{CVPR},
  2013.

\bibitem{OquabCVPR2014}
M.~Oquab, L.~Bottou, I.~Laptev, and J.~Sivic, ``Learning and transferring
  mid-level image representations using convolutional neural networks,'' in
  \emph{CVPR}, 2014.

\bibitem{SabzmeydaniCVPR2007}
P.~Sabzmeydani and G.~Mori, ``Detecting pedestrians by learning shapelet
  features,'' in \emph{CVPR}, 2007.

\bibitem{SinghECCV2012}
S.~Singh, A.~Gupta, and A.~Efros, ``Unsupervised discovery of mid-level
  discriminative patches,'' \emph{ECCV}, 2012.

\bibitem{SunICCV2013}
J.~Sun and J.~Ponce, ``Learning discriminative part detectors for image
  classification and cosegmentation,'' in \emph{ICCV}, 2013.

\bibitem{YaoICCV2011}
B.~Yao, X.~Jiang, A.~Khosla, A.~L. Lin, L.~J. Guibas, and L.~Fei-Fei, ``Action
  recognition by learning bases of action attributes and parts,'' in
  \emph{ICCV}, 2011.

\bibitem{PariziICLR2015}
S.~N. Parizi, A.~Vedaldi, A.~Zisserman, and P.~Felzenszwalb, ``Automatic
  discovery and optimization of parts for image classification,'' in
  \emph{ICLR}, 2015.

\bibitem{BoimanNIPS2006}
O.~Boiman and M.~Irani, ``Similarity by composition,'' in \emph{NIPS}, 2006.

\bibitem{LeibeIJCV2008}
B.~Leibe, A.~Leonardis, and B.~Schiele, ``Robust object detection with
  interleaved categorization and segmentation,'' \emph{IJCV}, vol.~77, no.~1,
  pp. 259--289, 2008.

\bibitem{BoimanCVPR2008}
O.~Boiman, E.~Shechtman, and M.~Irani, ``In defense of nearest-neighbor based
  image classification,'' in \emph{CVPR}, 2008.

\bibitem{ZhuECCV2012}
P.~Zhu, L.~Zhang, Q.~Hu, and S.~Shiu, ``Multi-scale patch based collaborative
  representation for face recognition with margin distribution optimization,''
  in \emph{ECCV}, 2012.

\bibitem{GuhaPAMI2012}
T.~Guha and R.~K. Ward, ``Learning sparse representations for human action
  recognition,'' \emph{PAMI}, vol.~34, no.~8, pp. 1576--1588, 2012.

\bibitem{AndrilukaCVPR2014}
M.~Andriluka, L.~Pishchulin, P.~Gehler, and B.~Schiele, ``{2D} human pose
  estimation: New benchmark and state of the art analysis,'' in \emph{CVPR},
  2014.

\bibitem{CharlesIJCV2014}
J.~Charles, T.~Pfister, M.~Everingham, and A.~Zisserman, ``Automatic and
  efficient human pose estimation for sign language videos,'' \emph{IJCV}, vol.
  110, no.~1, pp. 70--90, 2014.

\bibitem{DantonePAMI2014}
M.~Dantone, J.~Gall, C.~Leistner, and L.~Van~Gool, ``Body parts dependent joint
  regressors for human pose estimation in still images,'' \emph{PAMI}, vol.~36,
  no.~11, pp. 2131--2143, 2014.

\bibitem{FanCVPR2015}
X.~Fan, K.~Zheng, Y.~Lin, and S.~Wang, ``Combining local appearance and
  holistic view: Dual-source deep neural networks for human pose estimation,''
  in \emph{CVPR}, 2015.

\bibitem{TompsonNIPS2014}
J.~J. Tompson, A.~Jain, Y.~LeCun, and C.~Bregler, ``Joint training of a
  convolutional network and a graphical model for human pose estimation,'' in
  \emph{NIPS}, 2014.

\bibitem{ToshevCVPR2014}
A.~Toshev and C.~Szegedy, ``{DeepPose}: Human pose estimation via deep neural
  networks,'' in \emph{CVPR}, 2014.

\bibitem{VemulapalliCVPR2014}
R.~Vemulapalli, F.~Arrate, and R.~Chellappa, ``Human action recognition by
  representing {3D} skeletons as points in a lie group,'' in \emph{CVPR}, 2014.

\bibitem{ThurauCVPR2008}
C.~Thurau and V.~Hlavac, ``Pose primitive based human action recognition in
  videos or still images,'' in \emph{CVPR}, 2008.

\bibitem{ChenECCV2012}
H.~Chen, A.~Gallagher, and B.~Girod, ``Describing clothing by semantic
  attributes,'' in \emph{ECCV}, 2012.

\bibitem{YaoECCV2012}
B.~Yao and L.~Fei-Fei, ``Action recognition with exemplar based {2.5D} graph
  matching,'' in \emph{ECCV}, 2012.

\bibitem{ZhangCVPR2014}
N.~Zhang, M.~Paluri, M.~Ranzato, T.~Darrell, and L.~Bourdev, ``Panda: Pose
  aligned networks for deep attribute modeling,'' in \emph{CVPR}, 2014.

\bibitem{MaECCVW2012}
S.~Ma, S.~Sclaroff, and N.~Ikizler-Cinbis, ``Unsupervised learning of
  discriminative relative visual attributes,'' in \emph{ECCV Workshops}, 2012.

\bibitem{KumarPAMI2011}
N.~Kumar, A.~C. Berg, P.~N. Belhumeur, and S.~K. Nayar, ``Describable visual
  attributes for face verification and image search,'' \emph{PAMI}, vol.~33,
  no.~10, pp. 1962--1977, 2011.

\bibitem{RudovicPAMI2013}
O.~Rudovic, M.~Pantic, and I.~Patras, ``Coupled gaussian processes for
  pose-invariant facial expression recognition,'' \emph{PAMI}, vol.~35, no.~6,
  pp. 1357--1369, 2013.

\bibitem{SharmaECCV2012}
G.~Sharma, S.~ul~Hussain, and F.~Jurie, ``Local higher-order statistics ({LHS})
  for texture categorization and facial analysis,'' in \emph{ECCV}, 2012.

\bibitem{WanPR2014}
S.~Wan and J.~Aggarwal, ``Spontaneous facial expression recognition: A robust
  metric learning approach,'' \emph{PR}, vol.~47, no.~5, pp. 1859--1868, 2014.

\bibitem{LiCVPR2012}
C.~Li, Q.~Liu, J.~Liu, and H.~Lu, ``Learning ordinal discriminative features
  for age estimation,'' in \emph{CVPR}, 2012.

\bibitem{ChangTIP2015}
K.-Y. Chang and C.-S. Chen, ``A learning framework for age rank estimation
  based on face images with scattering transform,'' \emph{TIP}, vol.~24, no.~3,
  pp. 785--798, 2015.

\bibitem{GengPAMI2007}
X.~Geng, Z.-H. Zhou, and K.~Smith-Miles, ``Automatic age estimation based on
  facial aging patterns,'' \emph{PAMI}, vol.~29, no.~12, pp. 2234--2240, 2007.

\bibitem{GuoICCV2009}
G.~Guo, G.~Mu, Y.~Fu, C.~Dyer, and T.~Huang, ``A study on automatic age
  estimation using a large database,'' in \emph{ICCV}, 2009.

\bibitem{GuoCVPR2012}
G.~Guo and X.~Wang, ``A study on human age estimation under facial expression
  changes,'' in \emph{CVPR}, 2012.

\bibitem{ShaoICCV2013}
M.~Shao, L.~Li, and Y.~Fu, ``What do you do? occupation recognition in a photo
  via social context,'' in \emph{ICCV}, 2013.

\bibitem{GuoPR2014}
G.~Guo and A.~Lai, ``A survey on still image based human action recognition,''
  \emph{PR}, vol.~47, no.~10, pp. 3343--3361, 2014.

\bibitem{HussainBMVC2010}
S.~Hussain and B.~Triggs, ``Feature sets and dimensionality reduction for
  visual object detection,'' in \emph{BMVC}, 2010.

\bibitem{PerronninCVPR2012}
F.~Perronnin, Z.~Akata, Z.~Harchaoui, and C.~Schmid, ``Towards good practice in
  large-scale learning for image classification,'' in \emph{CVPR}, 2012.

\bibitem{OlivaIJCV2001}
A.~Oliva and A.~Torralba, ``Modeling the shape of the scene: A holistic
  representation of the spatial envelope,'' \emph{IJCV}, vol.~42, pp. 145--175,
  2001.

\bibitem{KrizhevskyNIPS2012}
A.~Krizhevsky, I.~Sutskever, and G.~E. Hinton, ``Imagenet classification with
  deep convolutional neural networks,'' in \emph{NIPS}, 2012.

\bibitem{PorikliCVPR2005}
F.~Porikli, ``Integral histogram: A fast way to extract histograms in cartesian
  spaces,'' in \emph{CVPR}, 2005.

\bibitem{CrowSIGGRAPH1984}
F.~C. Crow, ``Summed-area tables for texture mapping,'' \emph{SIGGRAPH},
  vol.~18, no.~3, pp. 207--212, 1984.

\bibitem{ViolaIJCV2001}
P.~Viola and M.~Jones, ``Robust real-time object detection,'' \emph{IJCV},
  vol.~4, pp. 51--52, 2001.

\bibitem{BayCVIU2008}
H.~Bay, A.~Ess, T.~Tuytelaars, and L.~V. Gool, ``{SURF}: {S}peeded up robust
  features,'' \emph{CVIU}, vol. 110, no.~3, pp. 346--359, 2008.

\bibitem{VekslerCVPR2003}
O.~Veksler, ``Fast variable window for stereo correspondence using integral
  images,'' in \emph{CVPR}, 2003.

\bibitem{AdamCVPR2006}
A.~Adam, E.~Rivlin, and I.~Shimshoni, ``Robust fragments-based tracking using
  the integral histogram,'' in \emph{CVPR}, 2006.

\bibitem{Vedaldi2008}
A.~Vedaldi and B.~Fulkerson, ``{VLFeat}: An open and portable library of
  computer vision algorithms,'' http://www.vlfeat.org/, 2008.

\bibitem{VedaldiCVPR2010}
A.~Vedaldi and A.~Zisserman, ``Efficient additive kernels using explicit
  feature maps,'' in \emph{CVPR}, 2010.

\bibitem{FanJMLR2008}
R.-E. Fan, K.-W. Chang, C.-J. Hsieh, X.-R. Wang, and C.-J. Lin, ``{LIBLINEAR}:
  A library for large linear classification,'' \emph{JMLR}, vol.~9, pp.
  1871--1874, 2008.

\bibitem{SimonyanICLR2015}
K.~Simonyan and A.~Zisserman, ``Very deep convolutional networks for
  large-scale image recognition,'' \emph{ICLR}, 2015.

\bibitem{SzegedyCVPR2015}
C.~Szegedy, W.~Liu, Y.~Jia, P.~Sermanet, S.~Reed, D.~Anguelov, D.~Erhan,
  V.~Vanhoucke, and A.~Rabinovich, ``Going deeper with convolutions,'' in
  \emph{CVPR}, 2015.

\bibitem{SermanetCVPR2013}
P.~Sermanet, K.~Kavukcuoglu, S.~Chintala, and Y.~LeCun, ``Pedestrian detection
  with unsupervised multi-stage feature learning,'' in \emph{CVPR}, 2013.

\bibitem{GirshickCVPR2014}
R.~Girshick, J.~Donahue, T.~Darrell, and J.~Malik, ``Rich feature hierarchies
  for accurate object detection and semantic segmentation,'' in \emph{CVPR},
  2014.

\bibitem{JiPAMI2013}
S.~Ji, W.~Xu, M.~Yang, and K.~Yu, ``{3D} convolutional neural networks for
  human action recognition,'' \emph{PAMI}, vol.~35, no.~1, pp. 221--231, 2013.

\bibitem{DengCVPR2009}
J.~Deng, W.~Dong, R.~Socher, L.-J. Li, K.~Li, and L.~Fei-Fei, ``Imagenet: {A}
  large-scale hierarchical image database,'' in \emph{CVPR}, 2009.

\bibitem{RazavianCVPR2014}
A.~S. Razavian, H.~Azizpour, J.~Sullivan, and S.~Carlsson, ``Cnn features
  off-the-shelf: an astounding baseline for recognition,'' in \emph{CVPR},
  2014.

\bibitem{Vedaldi2014}
A.~Vedaldi and K.~Lenc, ``Matconvnet -- convolutional neural networks for
  matlab,'' \emph{arXiv}, 2014.

\bibitem{LiNIPS2010}
L.~Li, H.~Su, E.~Xing, and L.~Fei-Fei, ``Object bank: A high-level image
  representation for scene classification and semantic feature
  sparsification,'' in \emph{NIPS}, 2010.

\bibitem{WangCVPR2010llc}
J.~Wang, J.~Yang, K.~Yu, F.~Lv, T.~Huang, and Y.~Gong, ``Locality-constrained
  linear coding for image classification,'' in \emph{CVPR}, 2010.

\end{thebibliography}

%
\begin{IEEEbiography} [{\includegraphics[width=1in,height=1.25in,clip,keepaspectratio]{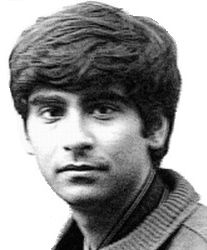}}]{Gaurav Sharma}
is currently with the Max Planck Institute for Informatics, Germany. He holds an Integrated M.Tech.\
(5 years programme) in Mathematics and Computing from the Indian Institute of Technology Delhi (IIT
Delhi), India and a PhD in Applied Computer Science from INRIA (LEAR team) and GREYC -- CNRS UMR6072,
University of Caen, France. His primary research interest lies in Computer Vision and applied
Machine Learning.
\end{IEEEbiography}

\begin{IEEEbiography}[{\includegraphics[width=1in,height=1.25in,clip,keepaspectratio]{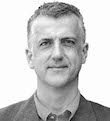}}]{Fr\'ed\'eric Jurie}
is a professor at the Universit\'e de Caen Basse-Normandie, France (GREYC -- CNRS UMR6072).  His
research interests lie predominately in the area of Computer Vision, particularly with respect to
object recognition, image classification and object detection.
\end{IEEEbiography}

\begin{IEEEbiography}[{\includegraphics[width=1in,height=1.25in,clip,keepaspectratio]{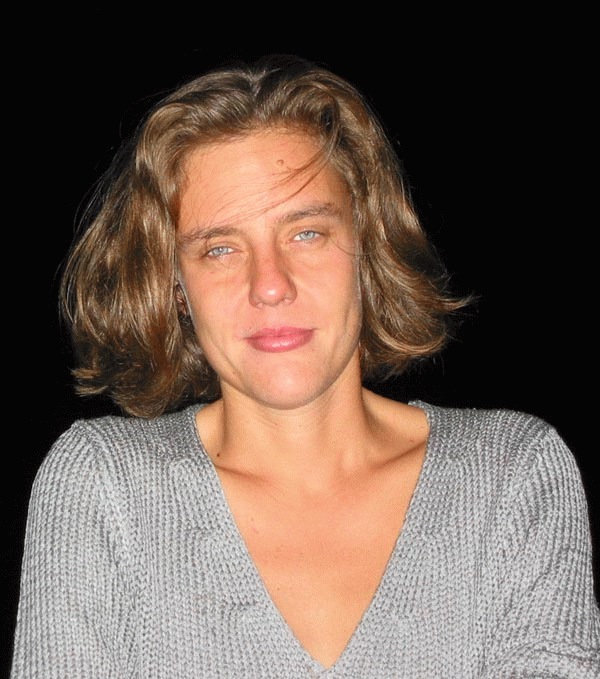}}]{Cordelia Schmid}
holds a M.S. degree from the University of Karlsruhe and a
Doctorate from the Institut National Polytechnique de Grenoble (INPG).  She is a research director
at INRIA, France, and directs the team LEAR. She is the author of over a hundred technical
publications. She was awarded the Longuet-Higgins prize for fundamental contributions in computer
vision that have withstood the test of time twice in 2006 \& 2014, an ERC advanced grant in
2012 and Humboldt prize in 2015. She is a fellow of IEEE. 
\end{IEEEbiography}




\end{document}